\documentclass[letterpaper, 10 pt, conference, onecolumn]{ieeeconf} 
\IEEEoverridecommandlockouts  
\usepackage{graphics} 
\usepackage{epsfig} 
\usepackage{mathptmx} 
\usepackage{amsmath} 
\usepackage{amssymb} 
\allowdisplaybreaks 
\usepackage{cite}
\usepackage{relsize} 
\usepackage[dvipsnames]{xcolor}
\usepackage{graphicx}
\usepackage{amsfonts}
\usepackage{mathtools}
\usepackage[mathcal]{euscript}
\usepackage[colorlinks=true,allcolors=steelblue]{hyperref}
\usepackage[T1]{fontenc} 
\usepackage{mathtools, cuted} 

\let\labelindent\relax 
\usepackage{enumitem} 
\usepackage{multirow, makecell}
\usepackage{diagbox}
\usepackage{balance} 
\usepackage{mathrsfs} 
\usepackage{soul} 
\usepackage{caption}
\usepackage{subcaption}

\usepackage[ruled, vlined, linesnumbered, noend]{algorithm2e}
\usepackage{algorithmic}

\DeclareCaptionLabelSeparator{periodspace}{.\quad}
\usepackage{amsthm}

\UseRawInputEncoding 

\newtheorem{remark}{Remark}

\theoremstyle{definition}

\newtheorem{definition}{Definition}
\definecolor{steelblue}{RGB}{70,130,180}

\def\bbr{\mathbb R}

\def\bbp{\mathbb P}
\def\bbe{\mathbb E}

\def\calx{\mathcal X}

\def\calc{\mathcal C}

\def\calo{\mathcal O}
\def\calt{\mathcal T}
\def\scrp{\mathscr P}
\def\scrt{\mathscr T}

\newcommand{\EE}{\mathbb{E}}
\newcommand{\RR}{\mathbb{R}}

\newcommand{\tp}{\intercal}

\DeclareMathOperator{\Tr}{Tr}

\newcommand{\RRT}{RRT}
\newcommand{\RRTs}{RRT*}
\newcommand{\ransrrt}{RANS-RRT*}

\newcommand{\absval}[1]{\mid {#1} \mid}

\newcommand{\norm}[1]{\left\lVert {#1} \right\rVert}
\newcommand{\rlbrack}[1]{\left [ {#1} \right]}
\newcommand{\rlbrace}[1]{\left \{ {#1}  \right\}}
\newcommand{\rlpar}[1]{\left ( {#1} \right)}
\newcommand{\mean}[1]{\overline{{#1}}}
\newcommand{\expect}[1]{\bbe \rlbrack{{#1}}}

\title{
Risk-Averse \RRTs{} Planning with Nonlinear Steering and Tracking Controllers for Nonlinear Robotic Systems Under Uncertainty \\
}

\author{\large Sleiman Safaoui*, Benjamin J. Gravell*, Venkatraman Renganathan*, Tyler H. Summers
\thanks{This work was supported in part by the Army Research Office under Grant W911NF-17-1-0058, in part by the United States Air Force Office of Scientific Research under Grant FA2386-19-1-4073, and in part by the National Science Foundation under Grant ECCS-2047040.}
\thanks{*Equal contribution of these authors. S. Safaoui is with the department of Electrical Engineering, and B. J. Gravell, V. Renganathan, and T. H. Summers are with the department of Mechanical Engineering at The University of Texas at Dallas, Richardson, TX, USA. E-mail: \{sleiman.safaoui, benjamin.gravell, vrengana, tyler.summers\}\@utdallas.edu.
}
}

\begin{document}

\maketitle
\thispagestyle{plain}
\pagestyle{plain}

\begin{abstract}
We propose a two-phase risk-averse architecture for controlling stochastic nonlinear robotic systems. We present Risk-Averse Nonlinear Steering \RRTs{} (\ransrrt{}) as an \RRTs{} variant that incorporates nonlinear dynamics by solving a nonlinear program (NLP) and accounts for risk by approximating the state distribution and performing a distributionally robust (DR) collision check to promote safe planning.  
The generated plan is used as a reference for a low-level tracking controller. We demonstrate three controllers: finite horizon linear quadratic regulator (LQR) with linearized dynamics around the reference trajectory, LQR with robustness-promoting multiplicative noise terms, and a nonlinear model predictive control law (NMPC).
We demonstrate the effectiveness of our algorithm using unicycle dynamics under heavy-tailed Laplace process noise in a cluttered environment. 
\end{abstract}

\section*{Supplementary Material}
Code supporting this project is available at \url{https://github.com/TSummersLab/RANS-RRTStar}.

\section{Introduction}\label{sec:introduction}
Safe deployment of mobile robots in uncertain dynamic environments, such as urban streets and crowded airspaces, requires a systematic accounting of various risks, both within and across layers in an autonomy stack. These autonomy stacks are naturally partitioned into a hierarchy of i) a high-level planner which generates a reference trajectory (often) offline before system operation, and ii) a low-level controller whose purpose is to track the reference trajectory in an online fashion and incorporate feedback to mitigate the effect of disturbances. The survey \cite{paden2016survey} examines several approaches for motion planning and control of autonomous ground vehicles and suggests two additional upper layers in the hierarchy, namely route planning and behavioral decision-making. In this paper, we assume such route plans and behavioral decisions are encapsulated by the motion planning and control problems.

Many motion planning algorithms have been developed under deterministic settings and assume linear robot dynamics in order to simplify their analysis and design. However, in practice, robotic systems are inherently both nonlinear and stochastic in nature due to external disturbances and noisy onboard sensors. In the presence of model uncertainty or process noise, the resulting trajectory is only a nominal reference and there are no guarantees of its safety. To account for the stochastic components and to provide probabilistic guarantees, motion planning under uncertainty has been considered in several lines of recent research \cite{blackmore2011chance, zhu2019chance}. Specifically, risk-aware motion planning algorithms for linear robot dynamics were developed recently using CVaR- \cite{hakobyan2019risk} and Wasserstein metric- \cite{hakobyan2020ICRA} based formulations. 

A chance-constrained version of \RRT{} and \RRTs{} respectively were proposed in \cite{luders2010chance, luders2013robust}, where chance constraints were used to encode the risk of constraint violation to provide probabilistic feasibility guarantees for robots with linear dynamics under additive uncertainties. On the other hand, these approaches made questionable assumptions of Gaussianity for system uncertainties ostensibly to maintain computational tractability. It was shown in \cite{summers2018distributionally} that such assumptions can lead to significant miscalculations of risk, and hence moment-based ambiguity sets were formulated to propose a distributionally robust variant of \RRT{} called DR-\RRT{}. This approach was extended in \cite{renganathan2020towards} to design an asymptotically optimal \RRTs{} using output feedback with linear quadratic regulator and Kalman filter-based state estimation. Here we take a first step towards designing risk-aware nonlinear steering-based motion plans for nonlinear robotic systems. This is closely aligned with the problem addressed by authors in \cite{insoon_nonlin_mp}. 
A low-level tracking controller is implemented to successfully track a given reference trajectory in the presence of uncertain process disturbances.  In this work, we assume perfect state estimates are available and consider full-state feedback controllers. The linear-quadratic regulator (LQR) controller being the most common can be obtained through dynamic programming where a quadratic cost involving state deviation and control effort is minimized. The LQR controller can further be generalized to achieve robust stability under parametric model uncertainties by designing it to mean-square stabilize the system with the inclusion of multiplicative noises as described in \cite{gravell2020ifac} (LQRm). However, both LQR controllers cannot handle state and input constraints. On the other hand, NMPC explicitly considers both state and input constraints \cite{magni_nmpc}. The authors in \cite{doerr2020motion} used the nonlinear model-predictive control (NMPC) to track the LQR-\RRTs{} trajectory to make up for linearization error. By contrast, we use NMPC to track a risk-averse \RRTs{} trajectory generated by a nonlinear program (NLP)-based steering function which much more closely resembles the low-level NMPC controller.

\textbf{Contributions:} 
\begin{enumerate}
    \item We present \ransrrt{}, a new sampling-based motion planner for nonlinear robotic systems which constructs dynamically feasible trajectories that satisfy distributionally robust state constraints to promote safety.
    \item We demonstrate our proposed approach on unicycle dynamics under heavy-tailed Laplace process noise in a cluttered environment. We provide a comparative study of the collision-avoidance rate, state deviation and control costs, and computational expense of three low-level reference tracking controllers i) LQR, ii) LQRm and iii) NMPC, across a range of disturbance strengths, through Monte Carlo simulations.
\end{enumerate}

The rest of the paper is organized as follows. The notation, preliminaries and problem formulations 
are presented in \S \ref{sec:prob_formulation}. The proposed nonlinear dynamics-based high level motion planner is elucidated in \S \ref{sec:hl_plan}. Low-level tracking controllers are described in \S \ref{sec:ll_plan}. Simulation results are reported and analyzed in \S \ref{sec:num_results}. Finally, the paper is closed in \S \ref{sec:conc} along with directions for future research.

\section*{Notations, Preliminaries \& Problem Formulation}  \label{sec:prob_formulation}
The set of real numbers and natural numbers are denoted by $\mathbb{R}, \mathbb{N}$ respectively. The subset of natural numbers between $a,b \in \mathbb{N}$ with $a < b$ is denoted by $[a:b]$. The operator $\backslash$ denotes set subtraction and $\absval{C}$ denotes the cardinality of the set $C$. An identity matrix in dimension $n$ is denoted by $I_{n}$. The operator $(\cdot)^c$ denotes the set complement.

\subsection{Environment and Obstacles Specification}
Consider a robot in an environment $\calx \subseteq \mathbb{R}^{n}$ with static obstacles. It is expected to navigate the environment $\mathcal{X}$ while safely avoiding obstacles at all times. We denote the set of all obstacles by $\mathcal{B}$ with $|\mathcal{B}| = F > 0$. 
The environment and obstacles (assumed disjoint) are convex polytopes and hence can each be represented as a conjunction of halfspace constraints. The space occupied by the $i^{th}$ obstacle in $\mathcal{B}$ is denoted $\mathcal{O}_{i}$. The union of the space occupied by all obstacles is $\calc := \cup_{i=1}^F\calo_i$. Hence the free space in the environment is given by
\begin{align*}
    \mathcal{X}_{free} &:= \ \calx \backslash \calc = \mathcal{X} \, \backslash \, \bigcup^{F}_{i = 1} \mathcal{O}_{i}  \\
    \calx &:= \rlbrace{x \mid A_{env} x \leq b_{env}} \\
    \calo_i &:= \rlbrace{x \mid A_{ob_i}x \leq b_{ob_i}} \quad \forall \calo_i \in \mathcal{B}.
\end{align*}
For a given deterministic state $s \in \bbr^n$, the condition for collision avoidance with all obstacles is
\begin{align*}
    s \not\in \calc \Leftrightarrow & \wedge_{i = 1}^{F} \rlpar{s \not\in \calo_i}, 
\end{align*}
where each individual obstacle avoidance constraint can be expressed as
\begin{align*}    
    s \not\in \calo_i \Leftrightarrow & \neg \rlpar{A_{ob_i}s \leq b_{obs_i}} \\
    \Leftrightarrow & \vee_{j = 1}^{n_{ob_i}} \rlpar{a_{ob_i,j}^{\tp}s \geq b_{ob_i,j}},
\end{align*}
and the condition for collision avoidance with the environment bounds is
\begin{align*}    
    s \in \calx \Leftrightarrow & \wedge_{j = 1}^{n_{env}} \rlpar{a_{env,j}^{\tp}s \leq b_{env,j}},
\end{align*}
where $n_{ob_i}$ are the number of constraints for obstacle $\calo_i$ and $n_{env}$ are those of the environment. The total number of constraints is denoted $n_{total} = n_{env} + \sum_{i=1}^{F}n_{ob_i}$.

\subsection{Robot System Dynamics} \label{sec:sys_dyn}
For all time instances $k \in \mathbb{N}$, we model the robot as a discrete-time nonlinear dynamical system given by:
\begin{align}
    x[k+1] = f(x[k], u[k]) + w[k], \qquad x[0] = x_0, \label{eq:nonlin_dyn}
\end{align}
where $x,w \in \bbr^n,\ u \in \bbr^m$ are the system state, additive disturbance, and control input, respectively, at the time step indexed in the brackets, $x_0 \in \bbr^n$ is the initial state, and $f: \mathbb{R}^{n} \times \mathbb{R}^{m} \rightarrow \mathbb{R}^{n}$ is the robot dynamics that represents the nonlinear transformation. 
The disturbances $w[k]$ are assumed independent and identically distributed according to some prescribed distribution $\bbp^w_{k} \sim (0, \Sigma^w_k)$. 

\subsection{Unscented Transformation and Moment Estimation}\label{sec:ukf}
The unscented transformation (UT) can be used to estimate the statistics of a random variable which undergoes a nonlinear transformation. An ensemble of $2n + 1$ samples called sigma points are generated deterministically \cite{wan2000unscented} and propagated individually through the nonlinear transformation to yield an ensemble of transformed sigma points. The weighted statistics of the transformed sigma points approximate the statistics of the transformed random variable. Though parameters that generate the sigma points can be tailored for specific distributions, there is no pre-defined set of rules that work in general for all distributions. For a discrete-time nonlinear robot system as in \eqref{eq:nonlin_dyn}, we can use the UT to estimate the mean and covariance of the next state given the current one. To do so, the ensemble of $2n + 1$ sigma points are obtained as follows 
\begin{equation*} 
    \chi_{i}[k] = \begin{cases}\hat{x}[k-1], &i = 0, \\
    \hat{x}[k-1] + \rlpar{ \sqrt{(n + \lambda)\Sigma_{x}[k-1]}}_{i}, &i = [1:n] \\
    \hat{x}[k-1] - \rlpar{ \sqrt{(n + \lambda)\Sigma_{x}[k-1]} }_{i-n}, &i = [n+1:2n].
    \end{cases} 
\end{equation*}
where  $(\sqrt{(n + \lambda)\Sigma_{x}[k-1]})_i$ is the $i^{th}$ row or column of the matrix $\sqrt{(n + \lambda)\Sigma_{x}[k-1]}$ 
obtained through Cholesky decomposition. 
Then, the weights below are used to scale $\chi_{i}[k]$ in the estimation of the mean and covariance
\begin{align} \label{eqn_wts_ut}
    W^{(m)}_{0} &= \frac{\lambda}{n + \lambda}, W^{(c)}_{0} = \frac{\lambda}{n + \lambda} + 1 - \hat{\alpha}^2 + \hat{\beta}, \\
    W^{(m)}_{i} &= W^{(c)}_{i} = \frac{\lambda}{2(n + \lambda)}, i = [1:2n]. \nonumber
\end{align}
$\lambda = \hat{\alpha}^2 (n + \kappa) - n$ is a scaling parameter where $\hat{\alpha}, \hat{\beta}, \kappa$ are used to tune the unscented transformation. Usually $\hat{\beta} = 2$ is a good choice for Gaussian uncertainties, $\kappa = 3 - n$ is a good choice for $\kappa$, and $0 \leq \hat{\alpha} \leq 1$ is an appropriate choice for $\hat{\alpha}$, where a larger value for $\hat{\alpha}$ spreads the sigma points further from the mean.
Using the above-obtained sigma points and the weights defined in \eqref{eqn_wts_ut}, the estimated mean and covariance of the random variable $x[k]$ under the dynamics \eqref{eq:nonlin_dyn}, assuming Gaussian noise $w$, are computed as follows.
\begin{align*}
    \xi_{i}[k] &= f(\chi_{i}[k], u_i[k-1]), \quad i = [0:2n], \\
    \hat{x}[k] &\simeq \sum^{2n}_{i=0}W^{(m)}_{i} \xi_{i}[k], \\ 
    \hat{\Sigma}_{x}[k] &\simeq \sum^{2n}_{i=0}W^{(c)}_{i} (\xi_{i}[k] - \hat{x}[k])(\xi_{i}[k] - \hat{x}[k])^{\tp} + \Sigma^w_k.
\end{align*}

\subsection{Moment-Based Ambiguity Set for The State Distribution}

The state $x[k] \ \forall k \in \mathbb{N}_{>0}$ is a random vector.
The state $x[k-1]$, under input $u[k-1]$ and the noise $w[k-1]$, evolves to $x[k]$. Due to the difficulty in estimating the distribution of a random variable under a nonlinear transformation, we will assume that a state $x[k]$ belongs to an unknown distribution $\bbp_k^x$. Since we can estimate its first two moments, we can consider a moment-based ambiguity set $\scrp_k^x$ with the estimated moments. This will guarantee robustness to errors in propagating the state distribution due to the nonlinear dynamics. It can also provide robustness to moment estimation errors.
The mean and covariance we consider for the ambiguity set are the ones we estimate through the UT and thus we get the following estimate for the ambiguity set:
\begin{align*}
 \scrp^x_k = \{ \bbp^x_{k} \mid & \bbe[x[k]] = \hat{x}[k], \  \bbe[(x[k]-\hat{x}[k])(x[k]-\hat{x}[k])^\tp] = \hat{\Sigma}_{x}[k]\}.
\end{align*}
For Gaussian inputs, the above moment estimates from UT are accurate up to the third-order approximation and for the case of non-Gaussian, the approximations are accurate to at least the second-order as described in \cite{wan2000unscented}. 

\subsection{Nonlinear Motion Planning Problem} \label{sec:planning_pb}
The planning problem provides a high-level solution that can then be used by low-level tracking controllers. This plan can be computed offline and requires finding an optimal reference trajectory that satisfies the robot dynamics, state and input constraints, and risk constraints. In this work, we consider the following planning problem.
\begin{definition}[\emph{Optimal Risk-Based Path Planning Problem}] \label{def:opt_path_plan}
    Given an initial state $x[0] \in \calx$ and a set of final goal locations $X_{goal} \subset \calx$, find a measurable state-and-input-history-dependent control policy $\pi = [\pi[0], \dots \pi[T-1]]$ with $u[k] = \pi[k](x[0:k], u[0:k-1])$ that minimizes the finite-horizon additive cost function subject to constraints:
    \begin{align}
        J_{hl} = \min_{\pi} \ & 
        \sum_{k=0}^{T-1}
        g_{hl}[k]\rlpar{\expect{x[k]}, u[k])} +
        g_{hl}[T]\rlpar{\expect{x[T]}} \label{eq:objective} \\
        \text{s.t.} \ & \eqref{eq:nonlin_dyn} \nonumber \\
        & w[k] \sim \bbp^w_{k} \label{eq:disturbance_distribution}\\
        & u[k] \in \mathcal{U}[k], \label{eq:input_constraints} \\
        & \inf_{\mathbb{P}^{x}_{k} \in \scrp^{x}_{k}} \mathbb{P}^{x}_{k} \rlpar{x[k] \in \mathcal{X}_{free}} \geq 1 - \alpha_k, \label{eq:risk_constraints} \\
        & \expect{x[T]} \in \calx_{goal} \label{eq:goal_constraint}.
    \end{align}
    Here, $g_{hl}[k]\  \forall k \in [1:T]$ is a stage cost function, \eqref{eq:input_constraints} is the inputs constraint, \eqref{eq:risk_constraints} is the states risk constraint that ensures that the state, under the worst-case distribution, is in the free space with high probability specified through the stage risk bound $\alpha_k \in (0, 0.5]$, and \eqref{eq:goal_constraint} is the goal constraint requiring the final state to be in the goal region.  
\end{definition}

\begin{remark}
    The risk constraint \eqref{eq:risk_constraints} is infinite dimensional and generally non-convex which makes solving this problem a challenge. 
\end{remark}

To understand the stage risk constraints \eqref{eq:risk_constraints} in the context of trajectory safety, let $P^S$ denote the event that plan $P$ succeeds and $P^F$ be the complementary event (i.e. failure). 
Consider the specification that a plan succeeds with high probability $\bbp(P^S) \geq 1 - \beta, \ \beta \in [0,0.5]$ or equivalently that it fails with low probability $\bbp(P^F) \leq \beta$. Failure requires at least one stage risk constraints to be violated.
Using the fact that 
\begin{align*}
    & \inf_{\mathbb{P}^{x}_{k} \in \scrp^{x}_{k}} \mathbb{P}^{x}_{k} \rlpar{x[k] \in \mathcal{X}_{free}} \geq 1 - \alpha_k \\
    \Leftrightarrow & \sup_{\mathbb{P}^{x}_{k} \in \scrp^{x}_{k}} \mathbb{P}^{x}_{k} \rlpar{x[k] \not\in \mathcal{X}_{free}} \leq \alpha_{k}
\end{align*}
and applying Boole's law, the probability of the success event can be lower bounded as follows:
\begin{align*}
    \bbp(P^S) &= 1 - \bbp(P^F)
    = 1 - \mathbb{P} \rlpar{ \bigcup_{k=0}^Tx[k] \not\in \mathcal{X}_{free}} \\
    &\geq 1 - \sup_{\mathbb{P}^{x}_{k} \in \scrp^{x}_{k} \forall k}
    \mathbb{P} \rlpar{ \bigcup_{k=0}^Tx[k] \not\in \mathcal{X}_{free} \ \Big| \ x[k] \sim \mathbb{P}^{x}_{k}} \\
    &\geq 1 - \sum_{k=0}^T \sup_{\mathbb{P}^{x}_{k} \in \scrp^{x}_{k}} \mathbb{P}^{x}_{k} \rlpar{x[k] \not\in \mathcal{X}_{free}} \\
    &\geq 1 - \underbrace{\sum_{k=0}^T \alpha_k}_{:=\beta}
\end{align*}
If the stage risks $\forall k \in [0:T]$ are equal, meaning $\alpha_k = \alpha$, then $\beta = (T+1) \alpha$. Furthermore, if the stage risk $\alpha_k = \alpha$ is equally distributed over all $n_{total}$ constraints, then, the risk bound for a single constraint is $\alpha/n_{total}$ and the risk bound for a single obstacle $\calo_i$ (or the environment $\calx$) is $\alpha n_{ob_i}/n_{total}$ (or $\alpha n_{env}/n_{total}$) (this will be discussed more in \S \ref{sec:risk_treatment}).

\begin{remark}
    In general, $T$ may not be known ahead of time. For sampling-based planners, $T$ depends on the random nodes sampled. In such cases, an upper bound can be used $T_{max} \geq T$. 
    The choices of $T_{max}$, 
    the risk bound on the plan's failure $\beta$, 
    and the risk budget allocation across time steps and constraints
    are design parameters.
    Alternatively, it is also possible to build up the risk bounds from the individual constraints into a risk bound on the whole plan \cite{safaoui2020control}.
\end{remark}

\subsection{Nonlinear Reference Trajectory Tracking Problem}
Given a reference trajectory $\mean{x}[k]$ for $k \in [0:T]$ generated by the high-level motion planner, the reference tracking problem involves minimizing deviations of the state $x[k]$ from the reference state $\mean{x}[k]$ subject to the nonlinear dynamics and realizations of all system uncertainties. This problem is formally presented below.
\begin{definition}[\emph{Optimal reference trajectory tracking problem}] \label{def:ref_tracking}
Given reference trajectory $\mean{x}[k]$ for $k \in [0:T]$ such that $\mean{x}[0] = x[0]$ and $\mean{x}[T] \in X_{goal}$, find a measurable state-and-input-history-dependent control policy $\pi = [\pi[0], \dots \pi[T-1]]$ with $u[k] = \pi[k](x[0:k], u[0:k-1])$ that minimizes the finite-horizon additive cost function subject to constraints:
\begin{align*}
    J_{ll} = \min_{\pi} \quad & 
    \expect{\sum_{k=0}^{T-1} (g_{ll}[k](\mean{x}[k], x[k], u[k])) + g_{ll}[T](\mean{x}[T], x[T])}, \\
    \text{s.t.} \quad & \eqref{eq:nonlin_dyn}, \eqref{eq:disturbance_distribution}, \eqref{eq:input_constraints}
\end{align*}
where $g_{ll}[k] \forall k \in [1:T]$ is a stage cost function that penalizes the control effort and deviations of the robot state from those of the reference trajectory at each time step.
\end{definition}

\section{High Level Planner: \ransrrt{}}\label{sec:hl_plan}

The high-level planner finds an optimal (or approximately optimal) plan for the low-level controller to execute.
If the plan gets close to obstacles, tracking might fail due to process noise.
By incorporating uncertainty in the high-level planner, a more conservative, but safe, trajectory is designed.
In this work, we present an approximate solution to the problem in Definition \ref{def:opt_path_plan} using rapidly exploring random trees. We propose \ransrrt{}: a Risk-Averse, Nonlinear Steering \RRTs{} planner. 
Below, we discuss: 1) the NLP problem used to steer between tree nodes, 2) the mean and covariance propagation along such a trajectory segment, 3) the treatment of uncertainty and risk, 4) the \ransrrt{} algorithm, and 5) a trajectory shortening post-processing step.

\subsection{NLP Steering}
Our \ransrrt{} algorithm employs a nonlinear steering law to compute a trajectory $\scrt$ of length $N$ consisting of state and input pairs $\scrt = \rlbrace{(s_0, u_0), \dots, (s_N, u_N)}$ that drive an initial state $s_{init}$ to a final one $s_{des}$. The first state belongs to the \ransrrt{} tree $\calt$ and the other is either a sampled state or another tree state. NLP steering is defined below.
\begin{definition}[NLP Steering] \label{def:nlp_steering}
    Given an initial state $s_{init}$, a desired state $s_{des}$, and a steering horizon $N$, the NLP steering solution is the set of states $\rlbrace{s_0, \dots, s_N}$ and controls $\rlbrace{u_0, \dots, u_{N-1}}$ to the following deterministic optimization problem without any  risk constraints.
    \begin{align}
        J_{nlp} = \min_{u[0:N-1]} \quad  & \sum_{k = 0}^{N-1} u_k^{\tp} R[k] u_k \\ 
        \text{s.t} \qquad & s_0 = s_{init} \nonumber\\
        & s_N = s_{des} \nonumber\\ 
        & s[k+1] = f(s[k], u[k]). \nonumber
    \end{align}
\end{definition}

\subsection{Mean and Covariance Propagation}
NLP steering returns the trajectory $\scrt = \rlbrace{(s_0, u_0), \dots, (s_{N-1}, u_{N-1}), (s_N)}$. To use this in the \ransrrt{} tree $\calt$, as a trajectory between two nodes, we need the mean state, covariance of the state, and control law. Since the dynamics are nonlinear and the NMPC control policy is the solution of a constrained optimization problem without an explicit form as a function of the state,
it is extremely challenging to incorporate the NMPC feedback law into the \ransrrt{} trajectories.
As such, we use the open-loop controls of the NLP trajectory $\scrt$. As a byproduct, we also require the mean states in the \ransrrt{} trajectory to match the states of the NLP trajectory $\scrt$. With the mean states and inputs fixed according to $\scrt$, we only need to estimate the covariance at the mean states. To do so, we use the UT with the NLP controls and match the UT mean states with the NLP states. 
This returns covariances associated with every state that are used in enforcing risk bounds. 
Thus, for every computed NLP trajectory $\scrt$ we obtain a \ransrrt{} trajectory of the form $\text{traj}_k = \rlbrace{(\hat{x}[Nk], \hat{\Sigma}[Nk], u[Nk]), \dots, (\hat{x}[N(k+1)], \hat{\Sigma}[N(k+1)])}$ where $\hat{x}[Nk+i] = s_i$ and $u[NK+i] = u_i$ for all $i$.

\subsection{Risk Treatment}\label{sec:risk_treatment}
Consider a mean state and covariance pair in the \ransrrt{} trajectories $(\hat{x}[k], \hat{\Sigma}[k])$. The risk constraint associate with this time step has the form 
\begin{align*}
    &\inf_{\bbp_k^x \in \scrp_k^x} \bbp_k^x\rlpar{x[k] \in \calx_{free}} \geq 1-\alpha_k\\
    \Leftrightarrow &
    \sup_{\bbp_k^x \in \scrp_k^x} \bbp_k^x\rlpar{x[k] \not\in \calx_{free}} \leq \alpha_k  \\
    \Leftrightarrow &
    \sup_{\bbp_k^x \in \scrp_k^x} \bbp_k^x\rlpar{x[k] \in \calc \cup \calx^c} \leq \alpha_k.
\end{align*}
Since $\calc$ is a union of the obstacle sets, we use Boole's law to get:
\begin{align*}
    \sup_{\bbp_k^x \in \scrp_k^x} \bbp_k^x\rlpar{x[k] \in \calc \cup \calx^c} 
    \leq 
    \sum_{i=1}^F &\sup_{\bbp_k^x \in \scrp_k^x} \bbp_k^x\rlpar{x[k] \in \calo_i} + \sup_{\bbp_k^x \in \scrp_k^x} \bbp_k^x\rlpar{x[k] \in \calx^c}
\end{align*}
We allocate the risk bound equally among the constraints by setting the risk bound for being in $\calo_i$ to $\alpha_k n_{ob_i}/n_{total}$ and that of not being in the environment to $\alpha_k n_{env}/n_{total}$. Thus:
\begin{align*}
    \sum_{i=1}^F &\sup_{\bbp_k^x \in \scrp_k^x} \bbp_k^x\rlpar{x[k] \in \calo_i} + \sup_{\bbp_k^x \in \scrp_k^x} \bbp_k^x\rlpar{x[k] \in \calx^c} \leq \sum_{i=1}^F \alpha_k n_{ob_i}/n_{total} + \alpha_k n_{env}/n_{total} = \alpha_k
\end{align*}
and hence, the desired risk constraint is enforced.
Now we turn our attention to satisfying:
\begin{align}
    \sup_{\bbp_k^x \in \scrp_k^x} \bbp_k^x\rlpar{x[k] \in \calx^c} &\leq \alpha_k n_{env}/n_{total} \label{eq:new_risk_const_env}\\
    \sup_{\bbp_k^x \in \scrp_k^x} \bbp_k^x\rlpar{x[k] \in \calo_i} &\leq \alpha_k n_{ob_i}/n_{total} \ \ \forall i \label{eq:new_risk_const_obs}
\end{align}

Since $x[k] \in \calx \ \Leftrightarrow \ A_{env}x[k] \leq b_{env} \ \Leftrightarrow \ \wedge_{j=1}^{n_{env}} a^\tp_{env,j}x[k] \leq b_{env,j}$, then $x[k] \in \calx^c \ \Leftrightarrow \ x[k] \not\in \calx \ \Leftrightarrow \ \vee_{j=1}^{n_{env}} a^\tp_{env,j}x[k] > b_{env,j}$ 
we can apply Boole's to \eqref{eq:new_risk_const_env} and get
\begin{align*}
    \sum_{j=1}^{n_{env}}\sup_{\bbp_k^x \in \scrp_k^x} \bbp_k^x\rlpar{a^\tp_{env,j}x[k] > b_{env,j}} \leq \alpha_k n_{env}/n_{total}.
\end{align*}
Thus, bound \eqref{eq:new_risk_const_env} is satisfied if 
\begin{align*}
    \sup_{\bbp_k^x \in \scrp_k^x} \bbp_k^x\rlpar{a^\tp_{env,j}x[k] > b_{env,j}} \leq \alpha_k/n_{total} \quad \forall j \in [1:n_{env}].
\end{align*}

As for \eqref{eq:new_risk_const_obs}, we have the following:
\begin{align}
    \eqref{eq:new_risk_const_obs} \Leftrightarrow & \inf_{\bbp_k^x \in \scrp_k^x} \bbp_k^x\rlpar{x[k] \not\in \calo_i} \geq  1- \alpha_k n_{ob_i}/n_{total} \nonumber \\
    \Leftrightarrow & \inf_{\bbp_k^x \in \scrp_k^x} \bbp_k^x\rlpar{\vee_{j=1}^{n_{ob_i}} a^\tp_{ob_i,j} x > b_{ob_i,j}} \geq  1- \alpha_k n_{ob_i}/n_{total} \nonumber
\end{align}
and
\begin{align}
    \inf_{\bbp_k^x \in \scrp_k^x} \bbp_k^x\rlpar{\vee_{j=1}^{n_{ob_i}} a^\tp_{ob_i,j} x > b_{ob_i,j}} & \geq \inf_{\bbp_k^x \in \scrp_k^x} \max_{j} \bbp_k^x\rlpar{a^\tp_{ob_i,j} x > b_{ob_i,j}} \geq \inf_{\bbp_k^x \in \scrp_k^x} \bbp_k^x\rlpar{a^\tp_{ob_i,j} x > b_{ob_i,j}} \label{eq:bool_lower_bound} 
\end{align}
where \eqref{eq:bool_lower_bound} follows from the Fr\'echet-Boole lower bound.
We then have the following result:
\begin{align*}
    \inf_{\bbp_k^x \in \scrp_k^x} \bbp_k^x\rlpar{a^\tp_{ob_i,j}x[k] > b_{ob_i,j}} \geq 1-\alpha_k/n_{total} \ \forall j \Rightarrow \inf_{\bbp_k^x \in \scrp_k^x} \bbp_k^x\rlpar{x[k] \not\in \calo_i} 
    \geq 1-\alpha_k/n_{total} 
    \geq 1- \alpha_k n_{ob_i}/n_{total}
\end{align*}
which holds as $n_{ob_i} \geq 1$,
i.e. by enforcing the left-hand-side, the right-hand-side (desired constraint) is guaranteed.
Therefore, \eqref{eq:risk_constraints} is satisfied if
\begin{align*}
    \sup_{\bbp_k^x \in \scrp_k^x} \bbp_k^x\rlpar{a^\tp_{env,j}x[k] > b_{env,j}} &\leq \alpha_k/n_{total} \ \forall j
    \\
    \inf_{\bbp_k^x \in \scrp_k^x} \bbp_k^x\rlpar{a^\tp_{ob_i,j}x[k] > b_{ob_i,j}} &\geq 1-\alpha_k/n_{total} \ \forall i, j.
\end{align*}
Using \cite[Theorem 3.1]{dr_cc_lp} these are equivalent to:
\begin{align}
    a^\tp_{env,j}\hat{x}[k] & \leq b_{env,j} - \sqrt{\frac{1-\alpha_k/n_{total}}{\alpha_k/n_{total}}} \norm{\hat{\Sigma}[k]^{1/2}a_{env,j}}_2 \ \forall j \label{eq:env_risk_tightened}
    \\
    a^\tp_{ob_i,j}\hat{x}[k] & \geq b_{ob_i,j} + \sqrt{\frac{1-\alpha_k/n_{total}}{\alpha_k/n_{total}}} \norm{\hat{\Sigma}[k]^{1/2}a_{ob_i,j}}_2 \ \forall i, j. \label{eq:obs_risk_tightened} 
\end{align}

\subsection{Algorithm}

\ransrrt{} is presented in Algorithm \ref{alg:RRT}. The free space is first sampled and the tree is initialized with the initial state (the root of the tree) and the initial covariance of zero (lines \ref{line:sample_space}-\ref{line:init_tree}).
Then the sampled states $x_{samples}$ are traversed. For every state $S \in x_{samples}$, the nearest node in the tree $S_{nearest} \in \calt$ is found (line \ref{line:find_nearest}). If the distance between $S$ and $S_{nearest}$ is larger than some threshold, a closer state $S_{lim}$ is returned along the same direction (line \ref{line:saturate}). In line \ref{line:steer}, NLP steering is performed to find a trajectory from $S_{nearest}$ to $S_{lim}$ within the steering horizon $N$. If steering fails, the sample is skipped. Else, a \ransrrt{} trajectory (mean states, covariances, and inputs) is returned. In line \ref{line:dr_check}, the trajectory is checked for collisions. A collision is detected if any of the risk-tightened constraints (\eqref{eq:env_risk_tightened}, \eqref{eq:obs_risk_tightened}) are violated at any of the mean states, in which case the sample is skipped. If safe, the trajectory may be added to the tree after checking nearby nodes for a more optimal trajectory as done in \RRTs{} (line \ref{line:connect_min_cost}). Then, the trajectory is added to the tree (line \ref{line:add_node_traj}) and $\calt$ undergoes the \RRTs{} rewiring set (line \ref{line:rewire}). Note that the rewire step also uses the same steering and collision check functions described before. When an optimal trajectory is queried, the trajectory with the smallest total NLP steering cost (output cost of Definition \ref{def:nlp_steering}) is returned.

\begin{algorithm}[!htb]
\SetAlgoLined
\KwResult{RANS-\RRTs{} Tree $\calt$}
 $S_{samples}$ = sample($\calx_{free}$)\; \label{line:sample_space}
 $\calt = [(\hat{x}_0, \hat{\Sigma}_0 = 0)]$\; \label{line:init_tree}
 \For{$S \in x_{samples}$}{
  $S_{nearest}$ = nearestNode($S, \calt$)\; \label{line:find_nearest}
  $S_{lim}$ = limitDistance($S, S_{nearest}$)\; \label{line:saturate}
  $(success,\ traj)$ = steer($S_{nearest}, S_{lim}$)\; \label{line:steer}
  \uIf{not $success$}{
  Continue\;}
  $collision$ = checkDRCollision($traj$)\; \label{line:dr_check}
  \uIf{$collision$}{
  Continue\;}
  $traj$ = connectViaMinCostPath($traj, \ \calt$)\; \label{line:connect_min_cost}
  $\calt$.addTrajNode($traj$)\; \label{line:add_node_traj}
  $\calt$.rewire()\; \label{line:rewire}
 }
 \caption{RANS-\RRTs{} - Tree Expansion}  \label{alg:RRT}
\end{algorithm}

\begin{remark}
    \ransrrt{} only \emph{approximately} solves the problem in Definition \ref{def:opt_path_plan} because 1) the risk treatment step \emph{approximately} solves the infinite-dimensional constraint \eqref{eq:risk_constraints}, 2) the covariance estimation step using UT is \emph{imperfect} for non-Gaussian distributions, and 3) the covariance propagation assumes the estimated means and NLP means \emph{coincide}, which is used to keep the problem tractable. 
    Hence, we do not make any formal risk-bound guarantees.
    Nonetheless, as we will see in the experimental results section, the padding added to obstacles through the DR collision check step makes the algorithm significantly robust to disturbances.
\end{remark}

\subsection{Post-Processing: Trajectory Shortening}
In the \ransrrt{} algorithm, the horizon $N$ used for solving the NLP steering problem is fixed. This is done to obtaining a decision on whether a trajectory exists between two nodes after solving one NLP problem. Ideally, we would want to solve the NLP steering problem with the shortest steering horizon. 
However, this is not trivial and might involve solving the steering problem with an increasing horizon after each failure (up to a certain upper bound). Since the NLP steering problem is computationally expensive, repeating the process would quickly make the problem intractable. 
To that end, we set the NLP steering horizon to a constant value $N := N_{hl}$ during \ransrrt{} but perform a post-processing trajectory-shortening step to the optimal trajectory. 

The post-processing step considers each trajectory $traj_k$ between two \ransrrt{} nodes and obtains a new NLP steering trajectory by solving the problem in Definition \ref{def:nlp_steering} for a different steering horizon $N$. $N$ is initially set to a small estimated value (which we assign based on the trajectory length and the robot dynamics). If NLP steering fails, $N$ is incremented $N = N+1$ and the process is repeated while $N<N_{hl}$ (at which point the original trajectory is used).
If a trajectory is found, the same covariance propagation and DR collision-avoidance steps as done in \ransrrt{} are performed. If the DR checks fail, $N$ is incremented and the process continues. If they succeed, the \ransrrt{} trajectory is updated with the shorter one.

We observed significantly shorter trajectories after performing this step. Furthermore, since the trajectory time is fixed based on the trajectory steering horizon ($t = N\Delta t$ where $\Delta t$ is the discrete-time step), \ransrrt{} trajectories have fixed duration regardless of the distance between nodes (causing slow and fast trajectories) whereas the shortened trajectories have different times leading to smoother trajectories.

\section{Tracking Controllers}\label{sec:ll_plan}

The low-level problem in Definition \ref{def:ref_tracking} was specified as the evaluation criterion of performance of the tracking controllers. However, our tracking controllers do not exactly solve the low-level problem in Definition \ref{def:ref_tracking}, but instead approximately solve it using established control design methodologies.
Compared with Definition \ref{def:ref_tracking}, the proposed tracking controllers make the following approximations.
The LQR and LQRm controllers assume dynamics are linearized about the reference trajectory (so \eqref{eq:nonlin_dyn} only holds approximately), and the input constraint \eqref{eq:input_constraints} is ignored. LQRm attempts to mitigate the effects of linearization error by treating the model errors as multiplicative noise.
NMPC solves an optimal control problem more similar to the one in Definition \ref{def:ref_tracking}, but the finite horizon $T$ is generally shorter and the effect of the process noise \eqref{eq:disturbance_distribution} is ignored by replacing the stochastic dynamics \eqref{eq:nonlin_dyn} with their expectation. Furthermore, we chose to add a state constraint that requires the nominal state to be in the environment $\calx$.
Details of the control design techniques are given throughout the remainder of this section for completeness.

In all low-level controllers we use identical low-level cost functions. This facilitates a fair comparison between controllers, as the same objective is approximately optimized.
We use stage costs which are quadratic in the state deviation and input:
\begin{align*}
    g_{ll}[k](\mean{x}[k], x[k], u[k]) &= \delta_x[k]^\tp Q[k] \delta_x[k] + u[k]^\tp R u[k] \\
    g_{ll}[T](\mean{x}[T], x[T]) &= \delta_x[T]^\tp Q[T] \delta_x[T]
\end{align*}
where
$
    \delta_x[k] = x[k] - \mean{x}[k]
$
and the penalty matrices $Q[k]$ and $R[k]$ are symmetric positive definite for all $k$.

\subsection{Robot dynamics}
We consider the problem of navigating a robot with unicycle dynamics from an initial state to a final set of states.
The discrete-time unicycle nonlinear dynamics obtained through forward-Euler discretization of the corresponding continuous-time dynamics are given by:
\begin{align} \label{eq:dt_uni_dyn}
    p_x[k+1] &= p_x[k] + \cos(\theta[k]) \nu[k] \Delta t + w_x[k] \Delta t \nonumber \\
    p_y[k+1] &= p_y[k] + \sin(\theta[k]) \nu[k] \Delta t + w_y[k] \Delta t \\
    \theta[k+1] &= \theta[k] + \omega[k] \Delta t + w_\theta[k] \Delta t \nonumber
\end{align}
where $p_x[k],p_y[k] \in \bbr$ are the horizontal and vertical positions of the robot, $\theta[k] \in \bbr$ is the heading of the robot relative to the $x$-axis of an assigned world frame, $\nu[k], \omega[k] \in \bbr$ are the linear and angular velocity control inputs, and $w_x[k], w_y[k], w_\theta[k]$ are the disturbances affecting each state, all expressed at timestamp $k$. The quantity $\Delta t$ is the sampling time in seconds between any two timestamps $k, k+1$. 
With short-hand notations
\begin{align*}
    x[k] &= \begin{bmatrix} p_x[k] & p_y[k] & \theta[k] \end{bmatrix}^\tp \\
    u[k] &= \begin{bmatrix} \nu[k] & \omega[k] \end{bmatrix}^\tp \\
    w[k] &= \begin{bmatrix} w_x[k] & w_y[k] & w_{\theta}[k] \end{bmatrix}^\tp    
\end{align*}
the dynamics in \eqref{eq:dt_uni_dyn} and corresponding Jacobians can be written in the compact form as
\begin{align*}
    x[k+1] &= f(x[k], u[k]) + w[k], \\
\frac{\partial f}{\partial x} \Big|_{\mean{x}, \bar{u}} &= 
    \begin{bmatrix}
    1 & 0 & -\nu \sin(\theta) \Delta t \\
    0 & 1 & \nu \cos(\theta) \Delta t \\
    0 & 0 & 1
    \end{bmatrix}, \,
    \frac{\partial f}{\partial u} \Big|_{\mean{x}, \bar{u}} = 
    \begin{bmatrix}
    \cos(\theta) & 0  \\
    \sin(\theta) & 0  \\
    0 & \Delta t
    \end{bmatrix} .
\end{align*}

\subsection{Generalized linear-quadratic control} \label{sec:gen_lqr}
In this subsection we propose and derive the solution to a generalized linear-quadratic control problem (LQP), which will be useful in the sequel.
Consider the following finite-horizon stochastic dynamic game:
\begin{align}
    \underset{\{u[k]\}_{k=0}^{T-1}}{\text{minimize}} \ \underset{\{v[k]\}_{k=0}^{T-1}}{\text{maximize}} \quad & \EE \sum_{k=0}^{T} c[k] \label{eq:glqp} \\
    \text{subject to } \quad  
    & x[k+1] = f[k](x[k], u[k], v[k], w[k]) \nonumber 
\end{align}  
with linear dynamics and quadratic stage costs
\begin{align*}
    f[k](x, u, v, w) 
    &=
    \widetilde{A}[k] x[k] + \widetilde{B}[k] u[k] + \widetilde{C}[k] v[k] + E[k] w[k], \\
    c[k] = g[k](x, u, v, z) 
    &=
    \begin{bmatrix}
    x \\
    u \\
    v \\
    z
    \end{bmatrix}^\tp
    \begin{bmatrix}
    G_{xx}[k] & G_{xu}[k] & G_{xv}[k] & G_{xz}[k] \\
    G_{ux}[k] & G_{uu}[k] & G_{uv}[k] & G_{uz}[k] \\
    G_{vx}[k] & G_{vu}[k] & G_{vv}[k] & G_{vz}[k] \\
    G_{zx}[k] & G_{zu}[k] & G_{zv}[k] & G_{zz}[k]
    \end{bmatrix}    
    \begin{bmatrix}
    x \\
    u \\
    v \\
    z
    \end{bmatrix}, \\
    c[T] = g[T](x, z) 
    &=
    \begin{bmatrix}
    x \\
    z
    \end{bmatrix}^\tp
    \begin{bmatrix}
    G_{xx}[T] & G_{xz}[T] \\
    G_{zx}[T] & G_{zz}[T]
    \end{bmatrix}    
    \begin{bmatrix}
    x \\
    z
    \end{bmatrix},
\end{align*}
where
\begin{align*}
    \widetilde{A}[k] &= A[k] + \sum_{i=1}^{n_\alpha} \alpha_i[k] A_i[k] \\
    \widetilde{B}[k] &= B[k] + \sum_{i=1}^{n_\beta} \beta_i[k] B_i[k] \\
    \widetilde{C}[k] &= C[k] + \sum_{i=1}^{n_\gamma} \gamma_i[k] C_i[k]
\end{align*}
where the expectation is with respect to the random variables
\begin{align*}
    {\{w[k], \alpha[k], \beta[k], \gamma[k]\}_{k=0}^{T-1}}
\end{align*}
which are distributed as
\begin{align*}
    & w[k] \sim \mathcal{D}_w[k](0, W[k]), \nonumber \\
    & \alpha_i[k] \sim \mathcal{D}_{\alpha_i}[k](0, \sigma^2_{\alpha_i}[k]) \text{ for } i=1, \ldots, n_\alpha  \nonumber \\
    & \beta_i[k] \sim \mathcal{D}_{\beta_i}[k](0, \sigma^2_{\beta_i}[k]) \text{ for } i=1, \ldots, n_\beta \nonumber \\
    & \gamma_i[k] \sim \mathcal{D}_{\gamma_i}[k](0, \sigma^2_{\gamma_i}[k]) \text{ for } i=1, \ldots, n_\gamma \nonumber    
\end{align*}
where $\mathcal{D}(0, X)$ is any distribution with mean zero and covariance $X$, and the random variables $x[0], \{w[k], \alpha[k], \beta[k], \gamma[k]\}_{k=0}^{T-1}$ are assumed statistically independent.
The variables have the following meanings: $x[k] \in \RR^n$ is the state, $u[k] \in \RR^m$ is a control input, $v[k] \in \RR^c$ is an adversarial input, $z[k] \in \RR^p$ is an exogenous signal, and $w[k] \in \RR^d$ is a disturbance. 
This formulation allows the policies to be shaped by the exogenous signal $z$ and its interaction with the state $x$ and inputs $u$ and $v$; specifically, in the sequel we will choose $z$ to be the concatenated reference state and input trajectory. 
The stochastic system matrices $\widetilde{A}[k] \in \RR^{n \times n}$, $\widetilde{B}[k] \in \RR^{n \times m}$, and $\widetilde{C}[k] \in \RR^{n \times c}$ are decomposed into the mean system matrices $A[k] \in \RR^{n \times n}$, $B[k] \in \RR^{n \times m}$, and $C[k] \in \RR^{n \times c}$ and multiplicative noise terms, which are further decomposed into the pattern matrices $A_i[k]$, $B_i[k]$, $C_i[k]$ and the scalar mutually independent random variables $\alpha_i[k]$, $\beta_i[k]$, $\gamma_i[k]$ which are distributed according to distributions $\mathcal{D}_{\alpha_i}[k]$, $\mathcal{D}_{\beta_i}[k]$, $\mathcal{D}_{\gamma_i}[k]$ which have mean zero and variances $\sigma^2_{\alpha_i}[k]$, $\sigma^2_{\beta_i}[k]$, $\sigma^2_{\gamma_i}[k]$.
The matrices $E[k] \in \RR^{n \times d}$ specify how the additive disturbance $w[k]$ enters the dynamics.
The matrices $G[k] \in \RR^{(n+m+c+p) \times (n+m+c+p)}$ are assumed to satisfy block semidefiniteness conditions that make the stage costs $g[k](x[k], u[k], v[k], z[k])$ convex in the state $x[k]$, control input $u[k]$, and exogenous signal $z[k]$, and concave in the adversarial input $v[k]$. The matrices $G[k]$ specify quadratic cost weights for each state, control input, adversarial input, and exogenous signal.
Notice that the stage costs $g[k]$ and dynamics $f[k]$ in the LQP are different from those used in the high-level planner, i.e. $g_{hl}[k]$ and $f_{hl}[k]$.

Optimal policies can be computed from optimal cost functions via dynamic programming backwards in time.
The initial cost function is
\begin{align*}
    J[T](x[T]) 
    &= 
    g[T](x[T], z[T]) \\
    &=
    \begin{bmatrix}
    x[T] \\
    z[T]
    \end{bmatrix}^\tp
    \begin{bmatrix}
    G_{xx}[T] & G_{xz}[T] \\
    G_{zx}[T] & G_{zz}[T] \\
    \end{bmatrix}    
    \begin{bmatrix}
    x[T] \\
    z[T]
    \end{bmatrix} \\
    &=
    \begin{bmatrix}
    x[T] \\
    1
    \end{bmatrix}^\tp
    \begin{bmatrix}
    P[T] & q[T] \\
    q[T]^\tp & r[T] \\
    \end{bmatrix}    
    \begin{bmatrix}
    x[T] \\
    1
    \end{bmatrix}
\end{align*}
where
\begin{align*}
  P[T] &= G_{xx}[T] \\
  q[T] &= G_{xz}[T] z[T] \\
  r[T] &= z[T]^\tp G_{zz}[T] z[T] ,
\end{align*}
i.e. a convex quadratic polynomial of $x[T]$.
The subsequent cost functions are found via the dynamic programming equation:
\begin{align*}
    J[k](x[k]) &= \min_{u[t]} \left\{ \mathcal{Q}[k](x[k], u[k], v[k], z[k]) \right\}
\end{align*}
where $\mathcal{Q}[k]$ is the state-action cost function
\begin{align*}
    \mathcal{Q}[k](x, u, v, z) 
    &=
    \EE \left[ g[k](x, u, v, z) + J[k+1](f[k](x, u, v, w[k])) \right],
\end{align*}
where expectation is with respect to ${w[k], \alpha[k], \beta[k], \gamma[k]}$.
We will show that if the current cost function $J[k+1](x)$ is quadratic in $x$, i.e. has the form
\begin{align*}
    J[k+1](x)
    &=
    \begin{bmatrix}
    x \\
    1
    \end{bmatrix}^\tp
    \begin{bmatrix}
    P[k+1] & q[k+1] \\
    q[k+1]^\tp & r[k+1] \\
    \end{bmatrix}    
    \begin{bmatrix}
    x \\
    1
    \end{bmatrix},
\end{align*}
then the prior cost function $J[k](x)$ is also quadratic in $x$, i.e.
\begin{align*}
    J[k](x)
    &=
    \begin{bmatrix}
    x \\
    1
    \end{bmatrix}^\tp
    \begin{bmatrix}
    P[k] & q[k] \\
    q[k]^\tp & r[k] \\
    \end{bmatrix}    
    \begin{bmatrix}
    x \\
    1
    \end{bmatrix},
\end{align*}
and we will provide the explicit relation between $P[k]$, $q[k]$, $r[k]$ and $P[k+1]$, $q[k+1]$, $r[k+1]$, which will also be needed to compute the optimal policies. Then, arguing by induction and noting that by assumption the terminal cost function $J[T](x)$ is quadratic in $x$, all cost functions until $k=0$ are quadratic in $x$ with the same form.

We begin by evaluating the state-action cost function using the assumption that $J[k+1](x)$ is quadratic in $x$:
\begin{align*}
    \mathcal{Q}[k](x, u, v, z) 
    &=
    \EE_{w[k], \alpha[k], \beta[k], \gamma[k]} \left[ g[k](x, u, v, z) + J[k+1](f[k](x, u, v, w[k])) \right] \\
    &=
    \EE_{w[k], \alpha[k], \beta[k], \gamma[k]} \left( 
    \begin{bmatrix}
    x \\
    u \\
    v \\
    z
    \end{bmatrix}^\tp
    \begin{bmatrix}
    G_{xx}[k] & G_{xu}[k] & G_{xv}[k] & G_{xz}[k] \\
    G_{ux}[k] & G_{uu}[k] & G_{uv}[k] & G_{uz}[k] \\
    G_{vx}[k] & G_{vu}[k] & G_{vv}[k] & G_{vz}[k] \\
    G_{zx}[k] & G_{zu}[k] & G_{zv}[k] & G_{zz}[k]
    \end{bmatrix}    
    \begin{bmatrix}
    x \\
    u \\
    v \\
    z
    \end{bmatrix} \right. \\
    & \quad + \left.
    \begin{bmatrix}
    \widetilde{A}[k] x[k] + \widetilde{B}[k] u[k] + \widetilde{C}[k] v[k] + E[k] w[k] \\
    1
    \end{bmatrix}^\tp
    \begin{bmatrix}
    P[k+1] & q[k+1] \\
    q[k+1]^\tp & r[k+1] \\
    \end{bmatrix}    
    \begin{bmatrix}
    \widetilde{A}[k] x[k] + \widetilde{B}[k] u[k] + \widetilde{C}[k] v[k] + E[k] w[k] \\
    1
    \end{bmatrix}
    \right) .
\end{align*}
To ease notation, for the following development we will drop the indices $[k]$ and $[k+1]$, so
\begin{align*}
    \mathcal{Q}(x, u, v, z) 
    &=
    \EE_{w, \alpha, \beta, \gamma} \left( 
    \begin{bmatrix}
    x \\
    u \\
    v \\
    z
    \end{bmatrix}^\tp
    \begin{bmatrix}
    G_{xx} & G_{xu} & G_{xv} & G_{xz} \\
    G_{ux} & G_{uu} & G_{uv} & G_{uz} \\
    G_{vx} & G_{vu} & G_{vv} & G_{vz} \\
    G_{zx} & G_{zu} & G_{zv} & G_{zz}
    \end{bmatrix}    
    \begin{bmatrix}
    x \\
    u \\
    v \\
    z
    \end{bmatrix}
    + 
    \begin{bmatrix}
    \widetilde{A} x + \widetilde{B} u + \widetilde{C} v + E w \\
    1
    \end{bmatrix}^\tp
    \begin{bmatrix}
    P & q \\
    q^\tp & r \\
    \end{bmatrix}    
    \begin{bmatrix}
    \widetilde{A} x + \widetilde{B} u + \widetilde{C} v + E w \\
    1
    \end{bmatrix}
    \right) \\
    &=
    \EE_{w, \alpha, \beta, \gamma} \Big( 
      x^\tp G_{xx} x + x^\tp G_{xu} u + x^\tp G_{xv} v + x^\tp G_{xz} z 
    + u^\tp G_{ux} x + u^\tp G_{uu} u + u^\tp G_{xv} v + u^\tp G_{uz} z \\
    & \qquad + v^\tp G_{vx} x + v^\tp G_{vu} u + v^\tp G_{vv} v + v^\tp G_{vz} z 
    + z^\tp G_{zx} x + z^\tp G_{zu} u + z^\tp G_{xv} v + z^\tp G_{zz} z \\
    & \qquad 
    + x^\tp \widetilde{A}^\tp P \widetilde{A} x + x^\tp \widetilde{A}^\tp P \widetilde{B} u + x^\tp \widetilde{A}^\tp P \widetilde{C} v + x^\tp \widetilde{A}^\tp P E w
    + u^\tp \widetilde{B}^\tp P \widetilde{A} x + u^\tp \widetilde{B}^\tp P \widetilde{B} u + u^\tp \widetilde{B}^\tp P \widetilde{C} v + u^\tp \widetilde{B}^\tp P E w \\
    & \qquad 
    + v^\tp \widetilde{C}^\tp P \widetilde{A} x + v^\tp \widetilde{C}^\tp P \widetilde{B} u + v^\tp \widetilde{C}^\tp P \widetilde{C} v + v^\tp \widetilde{C}^\tp P E w
    + w^\tp E^\tp P \widetilde{A} x + w^\tp E^\tp P \widetilde{B} u +  w^\tp E^\tp P \widetilde{C} v + w^\tp E^\tp P E w \\
    & \qquad 
    + x^\tp A^\tp q + u^\tp B^\tp q + v^\tp C^\tp q + w^\tp E^\tp q
    + q^\tp A x + q^\tp B u + q^\tp C v + q^\tp E w 
    + r \Big) \\
    &=
      x^\tp G_{xx} x + x^\tp G_{xu} u + x^\tp G_{xv} v + x^\tp G_{xz} z 
    + u^\tp G_{ux} x + u^\tp G_{uu} u + u^\tp G_{xv} v + u^\tp G_{uz} z \\
    & \qquad + v^\tp G_{vx} x + v^\tp G_{vu} u + v^\tp G_{vv} v + v^\tp G_{vz} z 
    + z^\tp G_{zx} x + z^\tp G_{zu} u + z^\tp G_{xv} v + z^\tp G_{zz} z \\
    & \quad 
    + x^\tp (A^\tp P A + \sum_{i=1}^{n_\alpha} \sigma^2_{\alpha_i} A_i^\tp P A_i) x 
    + u^\tp (B^\tp P B + \sum_{i=1}^{n_\beta} \sigma^2_{\beta_i} B_i^\tp P B_i) u
    + v^\tp (C^\tp P C + \sum_{i=1}^{n_\gamma} \sigma^2_{\gamma_i} C_i^\tp P C_i) v \\
    & \quad
    + x^\tp A^\tp P B u + x^\tp A^\tp P C v + u^\tp B^\tp P A x + v^\tp C^\tp P A x + u^\tp B^\tp P C v + v^\tp C^\tp P B u
    + \Tr(E^\tp P E W) \\
    & \quad 
    + x^\tp A^\tp q + u^\tp B^\tp q + v^\tp C^\tp q 
    + q^\tp A x + q^\tp B u + q^\tp C v 
    + r ,
\end{align*}
where the last step follows by the zero mean and independence assumptions on the relevant random variables.
Since $\mathcal{Q}(x, u, v, z)$ is convex in $u$, the minimum with respect to $u$ is found when the gradient with respect to $u$ is equal to 0:
\begin{align*}
    0 = \nabla_u \mathcal{Q}(x, u, v, z) 
    &=
    2G_{ux} x + 2 G_{uu} u + 2 G_{uv} v + 2 G_{uz} z 
    + 2 B^\tp P A x + 2 (B^\tp P B + \sum_{i=1}^{n_\beta} \sigma^2_{\beta_i} B_i^\tp P B_i) u + 2 B^\tp P C v
    + 2 B^\tp q \\
    0 &=
    [(G_{ux} + B^\tp P A) x + (G_{uv} + B^\tp P C) v + G_{uz} z + B^\tp q ] + [G_{uu} + B^\tp P B + \sum_{i=1}^{n_\beta} \sigma^2_{\beta_i} B_i^\tp P B_i] u
\end{align*}    
rearranging,
\begin{align}    
    u &= -[G_{uu} + B^\tp P B + \sum_{i=1}^{n_\beta} \sigma^2_{\beta_i} B_i^\tp P B_i]^{-1} [(G_{ux} + B^\tp P A) x + (G_{uv} + B^\tp P C) v + G_{uz} z + B^\tp q ] \label{eq:u_base}
\end{align}

Likewise, since $\mathcal{Q}(x, u, v, z)$ is concave in $v$, the maximum with respect to $v$ is found when the gradient with respect to $v$ is equal to 0:
\begin{align*}
    0 = \nabla_u \mathcal{Q}(x, u, v, z) 
    &= 
    2G_{vx} x + 2 G_{vu} u + 2 G_{vv} v + 2 G_{vz} z 
    + 2 C^\tp P A x + 2 (C^\tp P C + \sum_{i=1}^{n_\gamma} \sigma^2_{\gamma_i} C_i^\tp P C_i) u + 2 C^\tp P B u
    + 2 C^\tp q \\
    0 &=
    [(G_{vx} + C^\tp P A) x + (G_{vu} + C^\tp P B) u + G_{vz} z + C^\tp q ] + [G_{vv} + C^\tp P C + \sum_{i=1}^{n_\gamma} \sigma^2_{\gamma_i} C_i^\tp P C_i] v
\end{align*} 
rearranging
\begin{align}
    v &= -[G_{vv} + C^\tp P C + \sum_{i=1}^{n_\gamma} \sigma^2_{\gamma_i} C_i^\tp P C_i]^{-1} [(G_{vx} + C^\tp P A) x + (G_{vu} + C^\tp P B) u + G_{vz} z + C^\tp q ] \label{eq:v_base}
\end{align}

Rewrite the expressions for $u$ and $v$ as
\begin{align}
    u &= - H_{uu}^{-1} (H_{ux} x + H_{uv} v + H_{uz} z + B^\tp q) \label{eq:u_base2} \\
    v &= - H_{vv}^{-1} (H_{vx} x + H_{vu} u + H_{vz} z + C^\tp q) \label{eq:v_base2}
\end{align}
where
\begin{align*}
    H
    =
    G + 
    \begin{bmatrix}
    A & B & C & 0
    \end{bmatrix}^\tp 
    P
    \begin{bmatrix}
    A & B & C & 0
    \end{bmatrix}
    +
    \text{blkdiag}(\sum_{i=1}^{n_\alpha} \sigma^2_{\alpha_i} A_i^\tp P A_i, \sum_{i=1}^{n_\beta} \sigma^2_{\beta_i} B_i^\tp P B_i, \sum_{i=1}^{n_\gamma} \sigma^2_{\gamma_i} C_i^\tp P C_i, 0)
\end{align*}

Now we decouple the expressions for $u$ and $v$. First substitute \eqref{eq:v_base2} into \eqref{eq:u_base2} to obtain
\begin{align*}
    -H_{uu} u 
    &=
    H_{ux} x - H_{uv} H_{vv}^{-1} (H_{vx} x + H_{vu} u + H_{vz} z + C^\tp q) + H_{uz} z + B^\tp q \\
    &=
    H_{ux} x - H_{uv} H_{vv}^{-1} H_{vx} x - H_{uv} H_{vv}^{-1} H_{vz} z + H_{uz} z - H_{uv} H_{vv}^{-1} C^\tp q + B^\tp q - H_{uv} H_{vv}^{-1} H_{vu} u \\
    &=
    (H_{ux} - H_{uv} H_{vv}^{-1} H_{vx}) x + (H_{uz} - H_{uv} H_{vv}^{-1} H_{vz}) z + (B + C H_{vv}^{-1} H_{vu})^\tp q - H_{uv} H_{vv}^{-1} H_{vu} u
\end{align*}
rearranging
\begin{align*}
    (-H_{uu} + H_{uv} H_{vv}^{-1} H_{vu}) u 
    =
    (H_{ux} - H_{uv} H_{vv}^{-1} H_{vx}) x + (H_{uz} - H_{uv} H_{vv}^{-1} H_{vz}) z + (B + C H_{vv}^{-1} H_{vu})^\tp q
\end{align*}
rearranging
\begin{align*}
    u 
    &= [-H_{uu} + H_{uv} H_{vv}^{-1} H_{vu}]^{-1} [(H_{ux} - H_{uv} H_{vv}^{-1} H_{vx}) x + (H_{uz} - H_{uv} H_{vv}^{-1} H_{vz}) z + (B + C H_{vv}^{-1} H_{vu})^\tp q] \\
    &= K_u x + L_u z + e_u \\
    &= K_u x + s_u
\end{align*}
where the gains are defined as
\begin{align*}
    K_u &= [-H_{uu} + H_{uv} H_{vv}^{-1} H_{vu}]^{-1} (H_{ux} - H_{uv} H_{vv}^{-1} H_{vx}), \\
    L_u &= [-H_{uu} + H_{uv} H_{vv}^{-1} H_{vu}]^{-1} (H_{uz} - H_{uv} H_{vv}^{-1} H_{vz}), \\
    e_u &= [-H_{uu} + H_{uv} H_{vv}^{-1} H_{vu}]^{-1} (B + C H_{vv}^{-1} H_{vu})^\tp q, \\
    s_u &= L_u z + e_u.
\end{align*}
An analogous calculation for $v$ yields
\begin{align*}
    v
    &= [-H_{vv} + H_{vu} H_{uu}^{-1} H_{uv}]^{-1} [(H_{vx} - H_{vu} H_{uu}^{-1} H_{ux}) x + (H_{vz} - H_{vu} H_{uu}^{-1} H_{uz}) z + (C + B H_{uu}^{-1} H_{uv})^\tp q] \\
    &= K_v x + L_v z + e_v \\
    &= K_v x + s_v
\end{align*}
where the gains are defined as
\begin{align*}
    K_v &= [-H_{vv} + H_{vu} H_{uu}^{-1} H_{uv}]^{-1} (H_{vx} - H_{vu} H_{uu}^{-1} H_{ux}), \\
    L_v &= [-H_{vv} + H_{vu} H_{uu}^{-1} H_{uv}]^{-1} (H_{vz} - H_{vu} H_{uu}^{-1} H_{uz}), \\
    e_v &= [-H_{vv} + H_{vu} H_{uu}^{-1} H_{uv}]^{-1} (C + B H_{uu}^{-1} H_{uv})^\tp q, \\
    s_v &= L_v z + e_v.
\end{align*}

Substituting the optimal input into the state-action cost function, we obtain the optimal cost function
\begin{align*}
    J[k](x) &=
      x^\tp G_{xx} x + x^\tp G_{xu} (K_u x + s_u) + x^\tp G_{xv} (K_v x + s_v) + x^\tp G_{xz} z \\
    & \quad + (K_u x + s_u)^\tp G_{ux} x + (K_u x + s_u)^\tp G_{uu} (K_u x + s_u) + (K_u x + s_u)^\tp G_{xv} (K_v x + s_v) + (K_u x + s_u)^\tp G_{uz} z \\
    & \quad + (K_v x + s_v)^\tp G_{vx} x + (K_v x + s_v)^\tp G_{vu} (K_u x + s_u) + (K_v x + s_v)^\tp G_{vv} (K_v x + s_v) + (K_v x + s_v)^\tp G_{vz} z \\
    & \quad + z^\tp G_{zx} x + z^\tp G_{zu} (K_u x + s_u) + z^\tp G_{xv} (K_v x + s_v) + z^\tp G_{zz} z \\
    & \quad 
    + x^\tp (A^\tp P A + \sum_{i=1}^{n_\alpha} \sigma^2_{\alpha_i} A_i^\tp P A_i) x 
    + (K_u x + s_u)^\tp (B^\tp P B + \sum_{i=1}^{n_\beta} \sigma^2_{\beta_i} B_i^\tp P B_i) (K_u x + s_u)
    + (K_v x + s_v)^\tp (C^\tp P C + \sum_{i=1}^{n_\gamma} \sigma^2_{\gamma_i} C_i^\tp P C_i) (K_v x + s_v) \\
    & \quad
    + x^\tp A^\tp P B (K_u x + s_u) + x^\tp A^\tp P C (K_v x + s_v) + (K_u x + s_u)^\tp B^\tp P A x + (K_v x + s_v)^\tp C^\tp P A x \\
    & \quad + (K_u x + s_u)^\tp B^\tp P C (K_v x + s_v) + (K_v x + s_v)^\tp C^\tp P B (K_u x + s_u)
    + \Tr(E^\tp P E W) \\
    & \quad 
    + x^\tp A^\tp q + (K_u x + s_u)^\tp B^\tp q + (K_v x + s_v)^\tp C^\tp q 
    + q^\tp A x + q^\tp B (K_u x + s_u) + q^\tp C (K_v x + s_v)
    + r.
\end{align*}
Grouping terms in quadratic, linear, and constant quantities of the state, we find indeed that $J[k](x)$ is a convex quadratic of the state, i.e.
\begin{align*}
    J[k](x)
    &=
    \begin{bmatrix}
    x \\
    1
    \end{bmatrix}^\tp
    \begin{bmatrix}
    P[k] & q[k] \\
    q[k]^\tp & r[k] \\
    \end{bmatrix}    
    \begin{bmatrix}
    x \\
    1
    \end{bmatrix},
\end{align*}
with
\begin{align*}
    P[k] &= 
    \begin{bmatrix}
    I \\ K_u \\ K_v
    \end{bmatrix}^\tp 
    \left(
    \begin{bmatrix}
    G_{xx} & G_{xu} & G_{xv} \\ 
    G_{ux} & G_{uu} & G_{uv} \\
    G_{vx} & G_{vu} & G_{vv} 
    \end{bmatrix}
    +
    \begin{bmatrix}
    A & B & C
    \end{bmatrix}^\tp 
    P 
    \begin{bmatrix}
    A & B & C
    \end{bmatrix}
    +
    \text{blkdiag}\left(\sum_{i=1}^{n_\alpha} \sigma^2_{\alpha_i} A_i^\tp P A_i, \sum_{i=1}^{n_\beta} \sigma^2_{\beta_i} B_i^\tp P B_i, \sum_{i=1}^{n_\gamma} \sigma^2_{\gamma_i} C_i^\tp P C_i)  \right)
    \right)
    \begin{bmatrix}
    I \\ K_u \\ K_v
    \end{bmatrix}, \\
    q[k] &=
    \begin{bmatrix}
    I \\ K_u \\ K_v
    \end{bmatrix}^\tp 
    \begin{bmatrix}
    G_{xu} & G_{xv} & G_{xz} \\ 
    G_{uu} & G_{uv} & G_{uz} \\
    G_{vu} & G_{vv} & G_{vz} 
    \end{bmatrix}   
    \begin{bmatrix}
    s_u \\ s_v \\ z
    \end{bmatrix}
    +
    \begin{bmatrix}
    B^\tp P A + (B^\tp P B + \sum_{i=1}^{n_\beta} \sigma^2_{\beta_i} B_i^\tp P B_i) K_u \\
    C^\tp P A + (C^\tp P C + \sum_{i=1}^{n_\gamma} \sigma^2_{\gamma_i} C_i^\tp P C_i) K_v \\
    A + B K_u + C K_v
    \end{bmatrix}^\tp     
    \begin{bmatrix}
    s_u \\ s_v \\ q
    \end{bmatrix}, \\
    r[k] &=
    \begin{bmatrix}
    s_u \\ s_v \\ z
    \end{bmatrix}^\tp
    \begin{bmatrix}
    G_{uu} & G_{uv} & G_{uz} \\ 
    G_{vu} & G_{vv} & G_{vz} \\ 
    G_{zu} & G_{zv} & G_{zz}
    \end{bmatrix}    
    \begin{bmatrix}
    s_u \\ s_v \\ z
    \end{bmatrix}
    +
    \begin{bmatrix}
    B s_u \\ C s_v \\ 1
    \end{bmatrix}^\tp
    \begin{bmatrix}
    P & P & q \\ 
    P & P & q \\
    q^\tp & q^\tp & r + \Tr(E^\tp P E W)
    \end{bmatrix}    
    \begin{bmatrix}
    B s_u \\ C s_v \\ 1
    \end{bmatrix}.
\end{align*}
Notice that the control policies do not depend on the scalar part of the cost $r[k]$ or the noise covariance $W[k]$, so it is not strictly necessary to compute $r[k]$ or require specification of $E[k]$, $W[k]$ for the purpose of control.

Thus, we have Algorithm \ref{algorithm:algo_glqp} to solve the generalized linear quadratic optimal control problem \eqref{eq:glqp}.
\begin{algorithm}
\caption{Dynamic programming solution to \eqref{eq:glqp}}
\begin{algorithmic}[1]
\label{algorithm:algo_glqp}
    \REQUIRE 
    {\renewcommand{\arraystretch}{1.2}
    \begin{tabular}{|r l|}
        \hline
        Dynamics matrix sequences & $\{A[k]\}_{k=0}^{T-1}$, $\{B[k]\}_{k=0}^{T-1}$, $\{C[k]\}_{k=0}^{T-1}$, $\{E[k]\}_{k=0}^{T-1}$ \\
        Multiplicative noise pattern matrices & $\{A_i[k]\}_{k=0}^{T-1}$, $\{B_i[k]\}_{k=0}^{T-1}$, $\{C_i[k]\}_{k=0}^{T-1}$ \\
        Multiplicative noise variances  & $\{\sigma^2_{\alpha_i}[k]\}_{k=0}^{T-1}$, $\{\sigma^2_{\beta_i}[k]\}_{k=0}^{T-1}$, $\{\sigma^2_{\gamma_i}[k]\}_{k=0}^{T-1}$ \\
        Additive noise covariance matrix sequence & $\{W[k]\}_{k=0}^{T-1}$ \\
        Penalty matrix sequence & $\{G[k]\}_{k=0}^{T}$ \\
        Exogenous signal & $\{z[k]\}_{k=0}^{T-1}$ \\
        \hline
    \end{tabular}} \ \\ \ \\
    \STATE Initialize optimal cost function:
    \begin{align*}
      P[T] &= G_{xx}[T], \\
      q[T] &= G_{xz}[T] z[T], \\
      r[T] &= z[T]^\tp G_{zz}[T] z[T]
    \end{align*}
    \vspace{-\baselineskip}
    \FOR{$k=T-1,\ldots 1, 0$}
        \STATE Drop indices by setting
        \begin{align*}
        \begin{array}{ccc}
            P = P[k+1], & q = q[k+1], & r = r[k+1] \\
            A = A[k], & B = B[k], & C = C[k], \\
            A_i = A_i[k], & B_i = B_i[k], & C_i = C_i[k] \\
            \sigma^2_{\alpha_i} = \sigma^2_{\alpha_i}[k], & \sigma^2_{\beta_i} = \sigma^2_{\beta_i}[k], & \sigma^2_{\gamma_i} = \sigma^2_{\gamma_i}[k], \\
            E = E[k], & W = W[k], & G = G[k] 
        \end{array}
        \end{align*}
        \STATE Compute the intermediate matrix
        \begin{align*}
            H
            =
            G + 
            \begin{bmatrix}
            A & B & C & 0
            \end{bmatrix}^\tp 
            P
            \begin{bmatrix}
            A & B & C & 0
            \end{bmatrix}
            +
            \text{blkdiag}\left(\sum_{i=1}^{n_\alpha} \sigma^2_{\alpha_i} A_i^\tp P A_i, \sum_{i=1}^{n_\beta} \sigma^2_{\beta_i} B_i^\tp P B_i, \sum_{i=1}^{n_\gamma} \sigma^2_{\gamma_i} C_i^\tp P C_i, 0\right)
        \end{align*}
        \STATE Compute optimal control input gains:
        \begin{align*}
            K_u[k] = K_u &= [-H_{uu} + H_{uv} H_{vv}^{-1} H_{vu}]^{-1} (H_{ux} - H_{uv} H_{vv}^{-1} H_{vx}), \\
            L_u[k] = L_u &= [-H_{uu} + H_{uv} H_{vv}^{-1} H_{vu}]^{-1} (H_{uz} - H_{uv} H_{vv}^{-1} H_{vz}), \\
            e_u[k] = e_u &= [-H_{uu} + H_{uv} H_{vv}^{-1} H_{vu}]^{-1} (B + C H_{vv}^{-1} H_{vu})^\tp q, \\
            s_u[k] = s_u &= L_u z + e_u.
        \end{align*}
        and optimal adversary input gains:
        \begin{align*}
            K_v[k] = K_v &= [-H_{vv} + H_{vu} H_{uu}^{-1} H_{uv}]^{-1} (H_{vx} - H_{vu} H_{uu}^{-1} H_{ux}), \\
            L_v[k] = L_v &= [-H_{vv} + H_{vu} H_{uu}^{-1} H_{uv}]^{-1} (H_{vz} - H_{vu} H_{uu}^{-1} H_{uz}), \\
            e_v[k] = e_v &= [-H_{vv} + H_{vu} H_{uu}^{-1} H_{uv}]^{-1} (C + B H_{uu}^{-1} H_{uv})^\tp q, \\
            s_v[k] = s_v &= L_v z + e_v.
        \end{align*}
        \STATE Update optimal costs:
        \begin{align*}
            P[k] &= 
            \begin{bmatrix}
            I \\ K_u \\ K_v
            \end{bmatrix}^\tp 
            \begin{bmatrix}
            H_{xx} & H_{xu} & H_{xv} \\ 
            H_{ux} & H_{uu} & H_{uv} \\
            H_{vx} & H_{vu} & H_{vv} 
            \end{bmatrix}
            \begin{bmatrix}
            I \\ K_u \\ K_v
            \end{bmatrix}, \\
            q[k] &=
            \begin{bmatrix}
            I \\ K_u \\ K_v
            \end{bmatrix}^\tp 
            \begin{bmatrix}
            G_{xu} & G_{xv} & G_{xz} \\ 
            G_{uu} & G_{uv} & G_{uz} \\
            G_{vu} & G_{vv} & G_{vz} 
            \end{bmatrix}   
            \begin{bmatrix}
            s_u \\ s_v \\ z
            \end{bmatrix}
            +
            \begin{bmatrix}
            B^\tp P A + (B^\tp P B + \sum_{i=1}^{n_\beta} \sigma^2_{\beta_i} B_i^\tp P B_i) K_u \\
            C^\tp P A + (C^\tp P C + \sum_{i=1}^{n_\gamma} \sigma^2_{\gamma_i} C_i^\tp P C_i) K_v \\
            A + B K_u + C K_v
            \end{bmatrix}^\tp     
            \begin{bmatrix}
            s_u \\ s_v \\ q
            \end{bmatrix}, \\
            r[k] &=
            \begin{bmatrix}
            s_u \\ s_v \\ z
            \end{bmatrix}^\tp
            \begin{bmatrix}
            G_{uu} & G_{uv} & G_{uz} \\ 
            G_{vu} & G_{vv} & G_{vz} \\ 
            G_{zu} & G_{zv} & G_{zz}
            \end{bmatrix}    
            \begin{bmatrix}
            s_u \\ s_v \\ z
            \end{bmatrix}
            +
            \begin{bmatrix}
            B s_u \\ C s_v \\ 1
            \end{bmatrix}^\tp
            \begin{bmatrix}
            P & P & q \\ 
            P & P & q \\
            q^\tp & q^\tp & r + \Tr(E^\tp P E W)
            \end{bmatrix}    
            \begin{bmatrix}
            B s_u \\ C s_v \\ 1
            \end{bmatrix}.
        \end{align*}
    \ENDFOR
    \ENSURE Optimal policy sequence $\{K_u[k], L_u[k], e_u[k]\}_{k=0}^{T-1}$, and cost sequence $\{P[k], q[k], r[k]\}_{k=0}^{T}$.
\end{algorithmic}
\end{algorithm}
At runtime, the solution from Algorithm \ref{algorithm:algo_glqp} is used to generate control inputs at each time $k$ according to
\begin{align*}
    u[k] = K_u[k]x[k] + L_u[k] z[k] + e_u[k].
\end{align*}
Notice that these control actions are the summation of a state-feedback term $K_u[k]x[k]$ and a feedforward term $L_u[k] z[k] + e_u[k]$ which can be computed before running the system. Notice also that the adversary gains $K_v[k]$, $L_v[k]$, $e_v[k]$ are not needed explicitly, but rather promote robustness of the controller implicitly via their effect on the control gains $K_u[k]$, $L_u[k]$, $e_u[k]$.

\subsection{Tracking with Linear Controller}
In order to apply optimal linear control, we linearize the discrete-time nonlinear dynamics about various operating points in the joint state-input space. From this section on, we notate $f = f_{hl}$.
A first-order Taylor series approximation of $f(x,u,w)$ evaluated at the operating point ($\bar{x}, \bar{u}, \bar{w}$) is 
\begin{align*}
    f(x, u, w) 
    &\approx 
    f(\bar{x}, \bar{u}) 
    + \frac{\partial f}{\partial x} \Big|_{\bar{x}, \bar{u}} (x - \bar{x}) 
    + \frac{\partial f}{\partial u} \Big|_{\bar{x}, \bar{u}} (u - \bar{u}) 
    + \frac{\partial f}{\partial w} \Big|_{\bar{x}, \bar{u}} (w - \bar{w}) .
\end{align*}
The low-level controller receives a reference trajectory, a sequence of states, inputs and disturbances, from the high-level planner, denoted as $\{(\mean{x}[k], \mean{u}[k], \mean{w}[k])\}_{k=0}^T$. 
The low-level controller will use these as the operating points about which to linearize the dynamics, i.e. will assume the dynamics
\begin{align}
    x[k+1] - f(\mean{x}[k], \mean{u}[k], \mean{w}[k])
    &= 
    \frac{\partial f}{\partial x} \Big|_{\mean{x}[k], \mean{u}[k], \mean{w}[k]} (x[k] - \mean{x}[k]) \nonumber \\
    & \ + \frac{\partial f}{\partial u} \Big|_{\mean{x}[k], \mean{u}[k], \mean{w}[k]} (u[k] - \mean{u}[k]) \nonumber \\
    & \ + \frac{\partial f}{\partial w} \Big|_{\mean{x}[k], \mean{u}[k], \mean{w}[k]} (w[k] - \mean{w}[k]). \label{eq:lin_dyn1}
\end{align}
We will also assume that the reference trajectory satisfies the nonlinear dynamics constraint, i.e.
\begin{align*}
  x_{\text{r}}[k+1] = f(\mean{x}[k], \mean{u}[k], \mean{w}[k]) \text{ for all } k=0, 1, \ldots, T.
\end{align*}
Using this assumption, the dynamics in \eqref{eq:lin_dyn1} simplify to
\begin{align*}
    x[k+1] - x_{\text{r}}[k+1] 
    &=
    \frac{\partial f}{\partial x} \Big|_{\mean{x}[k], \mean{u}[k], \mean{w}[k]} (x[k] - \mean{x}[k]) \\
    &\ + \frac{\partial f}{\partial u} \Big|_{\mean{x}[k], \mean{u}[k], \mean{w}[k]} (u[k] - \mean{u}[k]) \\
    &\ + \frac{\partial f}{\partial w} \Big|_{\mean{x}[k], \mean{u}[k], \mean{w}[k]} (w[k] - \mean{w}[k]),
\end{align*}
Defining the state-, input-, and disturbance-deviation variables
\begin{align*}
    \delta_x[k] &= x[k] - \mean{x}[k], \\
    \delta_u[k] &= u[k] - \mean{u}[k], \\
    \delta_w[k] &= w[k] - \mean{w}[k],
\end{align*}
the dynamics in \eqref{eq:lin_dyn1} can be rewritten as
\begin{align}
    \delta_x[k+1] = A[k] \delta_x[k] + B[k] \delta_u[k] + E[k] \delta_w[k], \label{eq:lin_dyn2}
\end{align}
with
\begin{align*}
    A[k] &= \frac{\partial f}{\partial x} \Big|_{\mean{x}[k], \mean{u}[k], \mean{w}[k]}, \\
    B[k] &= \frac{\partial f}{\partial u} \Big|_{\mean{x}[k], \mean{u}[k], \mean{w}[k]}, \\
    E[k] &= \frac{\partial f}{\partial w} \Big|_{\mean{x}[k], \mean{u}[k], \mean{w}[k]},
\end{align*}
which is clearly seen as linear time-varying dynamics in the deviation variables.

We now define general stage costs which are quadratic in the state, input, deviation of state from reference state, and deviation of input from reference input:
\begin{align}
    g[k](x[k], u[k])
    &=
    x[k]^\tp Q x[k] + u[k]^\tp R u[k] + \delta_x[k]^\tp Q_\delta[k] \delta_x[k] + \delta_u[k]^\tp R_\delta[k] \delta_u[k], \quad k=0,1,\ldots, T-1, \nonumber \\
    g[T](x[T])
    &=
    x[T]^\tp Q x[T] + \delta_x[T]^\tp Q_\delta[T] \delta_x[T] \label{eq:quad_stage_cost},
\end{align}
where $Q[k], R[k]$ are positive semidefinite matrices which penalize deviation of the state and input from the origin and $Q_\delta[k], R_\delta[k]$ are positive semidefinite matrices which penalize the deviation of the state and input from the reference. 
We provide equations in this level of generality, but practically the most meaningful thing to do is set $Q=0$ and $R_\delta=0$; the former because keeping the robot near the (absolute) origin in state-space is contrary to our goals, and the latter because we really only care about the total control effort expended, which can be smaller than that used by the reference open-loop input sequence in the event that serendipitous disturbance realizations drive the robot towards the reference states ``for free.''
Choosing the particular values for $Q_\delta$ and $R$ is a somewhat subjective task, but we will use diagonal and time-invariant matrices with ``reasonable'' values chosen that empirically give a good balance between reference tracking and control effort.
We also make the assumption that the disturbance-deviations are independent, zero-mean and have covariance $W[k]$, i.e.
\begin{align}
    \delta_w[k] \sim \mathcal{D}_w[k](0, W[k]). \label{eq:noise_assumption}
\end{align}
Using the linearized dynamics in \eqref{eq:lin_dyn2}, the time-additive quadratic stage-costs in \eqref{eq:quad_stage_cost}, and the disturbance distribution assumption in \eqref{eq:noise_assumption}, we obtain the linear-quadratic optimal tracking problem
\begin{align*}
    \text{minimize } \quad & \sum_{k=0}^{T-1} \left( x[k]^\tp Q x[k] + u[k]^\tp R u[k] + \delta_x[k]^\tp Q_\delta[k] \delta_x[k] + \delta_u[k]^\tp R_\delta[k] \delta_u[k] \right) + x[T]^\tp Q x[T] + \delta_x[T]^\tp Q_\delta[T] \delta_x[T] \\
    \text{subject to } \quad  & \delta_x[k+1] = A[k] \delta_x[k] + B[k] \delta_u[k] + E_k \delta_w[k] \\
    & \delta_w[k] \sim \mathcal{D}_w[k](0, W[k]).
\end{align*}

We now show how to bring this tracking problem into the form of the generalized linear quadratic problem in \eqref{eq:glqp}. 
First, it is clear that the dynamics and disturbance distribution have the required form already.
Next, rewrite the state and input in terms of the deviations and references as
\begin{align*}
    x[k] &= \delta_x[k] + \mean{x}[k], \\
    u[k] &= \delta_u[k] + \mean{u}[k],
\end{align*}
so the stage cost can be expressed as
\begin{align*}
    g[k](\delta_x[k], \delta_u[k])
    &=
    \delta_x[k]^\tp Q_\delta[k] \delta_x[k] + \delta_u[k]^\tp R_\delta[k] \delta_u[k] + (\delta_x[k] + \mean{x}[k])^\tp Q (\delta_x[k] + \mean{x}[k]) + (\delta_u[k] + \mean{u}[k])^\tp R (\delta_u[k] + \mean{u}[k]) \\
    &=
    \begin{bmatrix}
    \delta_x[k] \\
    \delta_u[k] \\
    \mean{x}[k] \\
    \mean{u}[k] 
    \end{bmatrix}^\tp
    \begin{bmatrix}
    Q_\delta + Q & 0 & Q & 0 \\
    0 & R_\delta + R & 0 & R \\
    Q & 0 & Q & 0 \\
    0 & R & 0 & R
    \end{bmatrix}     
    \begin{bmatrix}
    \delta_x[k] \\
    \delta_u[k] \\
    \mean{x}[k] \\
    \mean{u}[k] 
    \end{bmatrix}.
\end{align*}
Treating the concatenated reference state and input trajectory as an exogenous signal, i.e.
\begin{align*}
    z[k] 
    =
    \begin{bmatrix}
    \mean{x}[k] \\
    \mean{u}[k]
    \end{bmatrix},
\end{align*}
the stage cost can be expressed as
\begin{align*}
    g[k](\delta_x[k], \delta_u[k], z[k])
    &=
    \begin{bmatrix}
    \delta_x[k] \\
    \delta_u[k] \\
    z[k]
    \end{bmatrix}^\tp
    \begin{bmatrix}
    G_{\delta_x \delta_x}[k] & G_{\delta_x \delta_u}[k] & G_{\delta_x z}[k] \\
    G_{\delta_u \delta_x}[k] & G_{\delta_u \delta_u}[k] & G_{\delta_u z}[k] \\
    G_{z\delta_x}[k] & G_{z \delta_u}[k] & G_{zz}[k]
    \end{bmatrix}      
    \begin{bmatrix}
    \delta_x[k] \\
    \delta_u[k] \\
    z[k]
    \end{bmatrix},    
\end{align*}
where
\begin{alignat*}{3}
          G_{\delta_x \delta_x}[k] & = Q_\delta + Q \qquad
        & G_{\delta_x \delta_u}[k] & = 0 \qquad
        & G_{\delta_x z}[k]        & = \begin{bmatrix} Q & 0 \end{bmatrix} \\
          G_{\delta_u \delta_x}[k] & = 0 \qquad
        & G_{\delta_u \delta_u}[k] & = R_\delta + R  \qquad
        & G_{\delta_u z}[k]        & = \begin{bmatrix} 0 & R \end{bmatrix} \\
          G_{z\delta_x}[k]         & = \begin{bmatrix} Q \\ 0 \end{bmatrix} \qquad
        & G_{z \delta_u}[k]        & = \begin{bmatrix} 0 \\ R \end{bmatrix} \qquad
        & G_{zz}[k]                & = \begin{bmatrix} Q & 0 \\ 0 & R \end{bmatrix}.
\end{alignat*}

At this point we re-introduce the additional terms which promote robustness of the linearized controller, as discussed in Section \ref{sec:gen_lqr}.
We assume an additive adversary disturbance affects the state-deviation dynamics directly, as well as state-deviation- and input-deviation- and adversary-input-multiplicative noises. We will only consider a quadratic penalty on the adversary input characterized by symmetric positive definite matrix $S$. Thus, the robustified linear-quadratic tracking problem becomes
\begin{align*}
    \text{minimize } \quad & \sum_{k=0}^{T-1} \left( x[k]^\tp Q x[k] + u[k]^\tp R u[k] + \delta_x[k]^\tp Q_\delta[k] \delta_x[k] + \delta_u[k]^\tp R_\delta[k] \delta_u[k] - \delta_v[k]^\tp S_\delta[k] \delta_u[k] \right) + x[T]^\tp Q x[T] + \delta_x[T]^\tp Q_\delta[T] \delta_x[T] \\
    \text{subject to } \quad  & \delta_x[k+1] = \widetilde{A}[k] \delta_x[k] + \widetilde{B}[k] \delta_u[k] + \widetilde{C}[k] \delta_v[k]  + E_k \delta_w[k] \\
    & \delta_w[k] \sim \mathcal{N}(0, W[k]), \\
    & \alpha_i[k] \sim \mathcal{D}_{\alpha_i}[k](0, \sigma^2_{\alpha_i}[k]) \text{ for } i=1, \ldots, n_\alpha  \nonumber \\
    & \beta_i[k] \sim \mathcal{D}_{\beta_i}[k](0, \sigma^2_{\beta_i}[k]) \text{ for } i=1, \ldots, n_\beta \nonumber \\
    & \gamma_i[k] \sim \mathcal{D}_{\gamma_i}[k](0, \sigma^2_{\gamma_i}[k]) \text{ for } i=1, \ldots, n_\gamma \nonumber
\end{align*}
where
\begin{align*}
    \widetilde{A}[k] &= A[k] + \sum_{i=1}^{n_\alpha} \alpha_i[k] A_i[k] \\
    \widetilde{B}[k] &= B[k] + \sum_{i=1}^{n_\beta} \beta_i[k] B_i[k] \\
    \widetilde{C}[k] &= C[k] + \sum_{i=1}^{n_\gamma} \gamma_i[k] C_i[k]
\end{align*}
where
\begin{alignat*}{4}
          G_{\delta_x \delta_x}[k] & = Q_\delta + Q \qquad
        & G_{\delta_x \delta_u}[k] & = 0 \qquad
        & G_{\delta_x \delta_v}[k] & = 0 \qquad
        & G_{\delta_x z}[k]        & = \begin{bmatrix} Q & 0 \end{bmatrix} \\
          G_{\delta_u \delta_x}[k] & = 0 \qquad
        & G_{\delta_u \delta_u}[k] & = R_\delta + R  \qquad
        & G_{\delta_u \delta_v}[k] & = 0  \qquad
        & G_{\delta_u z}[k]        & = \begin{bmatrix} 0 & R \end{bmatrix} \\
          G_{\delta_v \delta_x}[k] & = 0 \qquad
        & G_{\delta_v \delta_u}[k] & = 0  \qquad
        & G_{\delta_v \delta_v}[k] & = -S_\delta  \qquad
        & G_{\delta_v z}[k]        & = \begin{bmatrix} 0 & 0 \end{bmatrix} \\   
          G_{z\delta_x}[k]         & = \begin{bmatrix} Q \\ 0 \end{bmatrix} \qquad
        & G_{z \delta_u}[k]        & = \begin{bmatrix} 0 \\ R \end{bmatrix} \qquad
        & G_{z \delta_v}[k]        & = \begin{bmatrix} 0 \\ 0 \end{bmatrix} \qquad
        & G_{zz}[k]                & = \begin{bmatrix} Q & 0 \\ 0 & R \end{bmatrix},
\end{alignat*}
which matches the required form of the stage costs in \eqref{eq:glqp}. Thus, we can apply Algorithm \ref{algorithm:algo_glqp} to compute the optimal policies.
At runtime, the control inputs are computed as
\begin{align*}
    u[k] 
    = 
    K_u[k] (x[k] - \mean{x}[k]) 
    + L_u[k] 
    \begin{bmatrix}
    \mean{x}[k] \\
    \mean{u}[k]
    \end{bmatrix}
    + e_u[k] 
    + \mean{u}[k],
\end{align*}
which can again be interpreted as the summation of a closed-loop feedback term of the state-deviation and an open-loop feedforward term.
As long as the state of the system remains close to the reference trajectory, the linearized dynamics will remain a good approximation and the linear controller will yield good reference tracking.

\subsection{LQR}\label{sec:lqr}
For vanilla LQR, we do not include any robustness-inducing terms, i.e. set $\alpha_i[k] = \beta_i[k] = \gamma_i[k] = 0$ and $C[k]=0$.

\subsection{Robust LQR}\label{sec:rob_lqr}
\begin{figure}[h]
    \centering
    \includegraphics[width=0.4\linewidth, trim={0cm 0cm 0cm 0cm},clip]{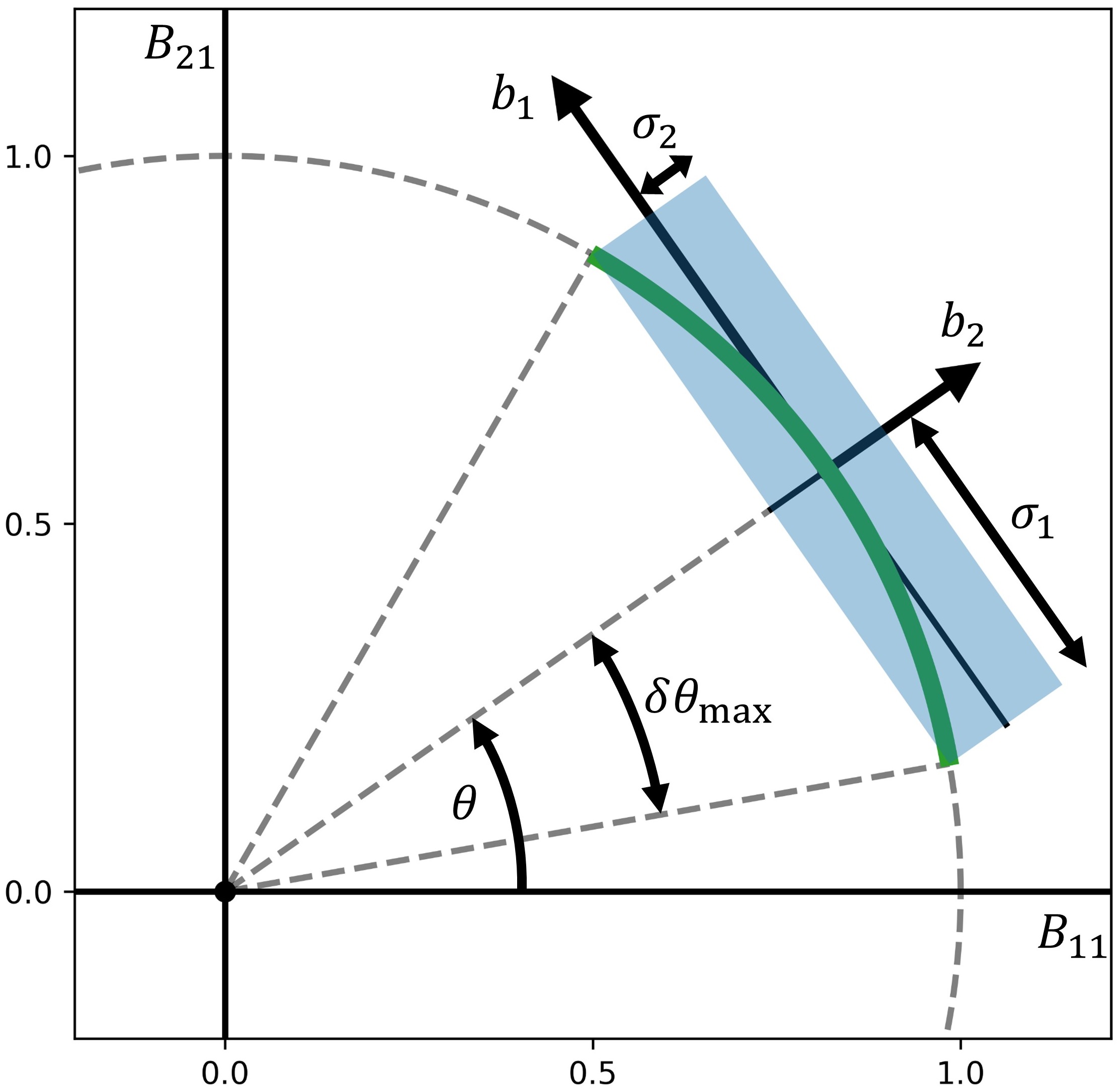}
    \caption{Geometry of $B$ matrices under linearization about various states. The thick circular arc segment is the locus of all possible $B$ when $\theta$ is interval bounded. The shaded box represents the set of $B$ on which the controller is designed to achieve low cost.}
    \label{fig:robustness_circle}
\end{figure}

We seek to promote robustness against errors in the state-space matrices due to linearization about states other than the reference trajectory, which occurs due to the process disturbance.
Observing the Jacobians, only mis-specifications in heading $\theta$ change the entries. We assume that the heading deviation from reference is uniformly upper bounded as
\begin{align*}
    |\theta[k] - \theta_{r}[k]| \leq \delta \theta_{\max} \leq \pi/2.
\end{align*}
For the  unicycle model, it is straightforward to construct appropriate 2D bounding boxes in the space of A and B matrices that fully contain every possible Jacobian.
Consider the $1,1$- and $2,1$-entries of the $B$ matrix
\begin{align*}
    b = 
    \begin{bmatrix}
    B_{11} \\
    B_{21}
    \end{bmatrix} 
    =
    \begin{bmatrix}
    \cos(\theta)  \\
    \sin(\theta)
    \end{bmatrix} .
\end{align*}
Through trigonometric relations depicted by Figure \ref{fig:robustness_circle}, we obtain the robustness regions 
\begin{align*}
    \overline{A}[k] &\in \{ A : \ A = A[k] + \mu_{A_1}[k] A_1[k] + \mu_{A_2}[k] A_2[k]\} \\
    \overline{B}[k] &\in \{ B : \ B = B[k] + \mu_{B_1}[k] B_1[k] + \mu_{B_2}[k] B_2[k]\}
\end{align*}
determined by directions
\begin{alignat*}{2}
    & A_1[k] = 
    \begin{bmatrix}
    0 & 0 & -s[k] \\
    0 & 0 & c[k] \\
    0 & 0 & 0
    \end{bmatrix}, 
    \
    && A_2[k] = 
    \begin{bmatrix}
    0 & 0 & -c[k] \\
    0 & 0 & -s[k]\\
    0 & 0 & 0
    \end{bmatrix},    \\
    & B_1[k] = 
    \begin{bmatrix}
    -s[k] & 0 \\
    c[k] & 0 \\
    0 & 0
    \end{bmatrix} ,
    \
    && B_2[k] = 
    \begin{bmatrix}
    c[k] & 0 \\
    s[k] & 0 \\
    0 & 0
    \end{bmatrix}    
\end{alignat*}
where $c[k] = \cos(\theta[k]), s[k] = \sin(\theta[k])$
and where the scales are bounded as 
\begin{align*}
    |\mu_{A_i}[k]| \leq \sigma_{A_i}[k] , \quad
    |\mu_{B_i}[k]| \leq \sigma_{B_i}[k]
\end{align*}
where
\begin{alignat*}{2}
    &  \sigma_{A_1}[k] = \nu[k] \Delta t \sin(\delta \theta_{\max}), \quad 
    && \sigma_{A_2}[k] = \nu[k] \Delta t \left[ 1 - \cos(\delta \theta_{\max}) \right], \\
    &  \sigma_{B_1}[k] = \sin(\delta \theta_{\max}), \quad
    && \sigma_{B_2}[k] = 1 - \cos(\delta \theta_{\max}).
\end{alignat*}

Note that the robustness region is conservative as it extends in the $A_2$ and $B_2$ directions twice as far as strictly necessary. This is done merely as a matter of convenience so that the center of the robustness region remains at ($A[k]$, $B[k]$) regardless of $\delta \theta_{\max}$. It has been proved in \cite{gravell2020ifac} that, in the infinite-horizon time-invariant setting, the inclusion of (fictitious) multiplicative noise in the control design induces robustness to static model perturbations in the same directions as the multiplicative noise. This inspires the inclusion of such multiplicative noises with variance and directions related to the desired robustness region over which we desire to minimize quadratic cost. 
We include the robust LQR in our comparison to demonstrate the benefit of our NMPC controller over a simpler robust control method.

\subsection{NMPC}\label{sec:nmpc}
NMPC can be used to approximate the nonlinear optimal trajectory tracking problem.
NMPC accomplishes that by reducing the problem into a sequence of open-loop optimization problems over a horizon $N_{ll}$ where after solving an NLP to track a reference trajectory, only the first input is applied and the horizon is shifted back. The NMPC tracking NLP is given below.
\begin{definition}[NMPC Tracking NLP]
Given an reference trajectory $\mean{x}[k]$ for $k \in [t:t+N_{ll}]$ and a planning horizon $N_{ll}$, find a control sequence at time step $t$ that minimizes the $N_{ll}$-horizon additive cost function subject to constraints:
    \begin{align*}
        \min_{u[t:t+N_{ll}-1]} \quad & 
        \sum_{k=t}^{t+N_{ll}-1} g_{ll}[k](\mean{x}[k], x[k], u[k]) + g_{ll}[T](\mean{x}[N_{ll}], x[N_{ll}])\\
        & \eqref{eq:input_constraints} \\
        & x[k] \in \calx \quad \forall k\\ 
        & x[k+1] = f(x[k], u[k]) \quad \forall k
    \end{align*}
\end{definition}
After solving this problem at step $t$, only the first input $u[t]$ is applied and the horizon is shifted to $t+1$ for the problem to be solved again. The final set of control inputs is thus $ u^*_{0:T-1} = [u[0], \dots u[T-1]]$.

\section{Numerical Results}\label{sec:num_results}
\ransrrt{} plans were generated on a machine with an Intel Core i7 6700K CPU and 16 of RAM. 
The low-level Monte Carlo simulations were performed on a machine with a Ryzen 7 2700X and 64GB of RAM. 
The NLP is modeled with CasADi Opti and solved with IPOPT \cite{Andersson2019}. 
We showcase \ransrrt{} in action in three different environments. Each environment consisted of a root node (white triangle), a goal area (dashed green rectangle), and a $10 \times 10$ environment with 4 rectangular obstacles for its sides (black boundary). Environments 1 and 3 (Figures \ref{fig:ransrrt_tree_1} and \ref{fig:ransrrt_tree_3}) have 5 rectangular obstacles (black rectangles) while environment 2 (Figure \ref{fig:ransrrt_tree_2}) has only 3 such obstacles.
The robot is assumed to occupy a single point. Its controls bounds are $\pm 0.5$ units/sec for linear velocity and $\pm \pi$rad/sec for angular velocity. The \ransrrt{} steering horizon is $N = N_{hl} = 30$ and the NMPC planning horizon is $N_{ll} = 10$. The discrete-time step was $\Delta t = 0.2$sec. The planner control cost matrix was $R[k] = \text{diag}([1,1])$. The tracking (LQR, robust LQR, and NMPC) cost matrices were $Q[k] = \text{diag}([100, 100, 10])$ and $R[k] = \text{diag}([1, 1])$ for all $k=[0 : T-1]$, and $Q[T] = 10 Q[0]$.

We used a high-level plan risk bound of $\beta = 0.1$. It is divided equally across the time steps $T_{max} = 1000$ and among the obstacle constraints. The process noise distribution $\bbp^w_{k}$ was taken as a multivariate Laplace distribution with zero mean and covariance $\Sigma^w = 5\cdot 10^{-7} I_{n}$. The same variance was used for all states and is denoted by $\sigma_w^2 := 5\cdot 10^{-7}$. We also refer to this as the noise level. In the tracking step, we evaluated performance under different (greater) noise levels. 

\subsection{\ransrrt{} Results}
\begin{figure}[p!]
    \centering
    \begin{subfigure}[t]{0.48\linewidth}
        \centering
        \includegraphics[width=1.0\linewidth,trim={3cm 2cm 2cm 2cm},clip]{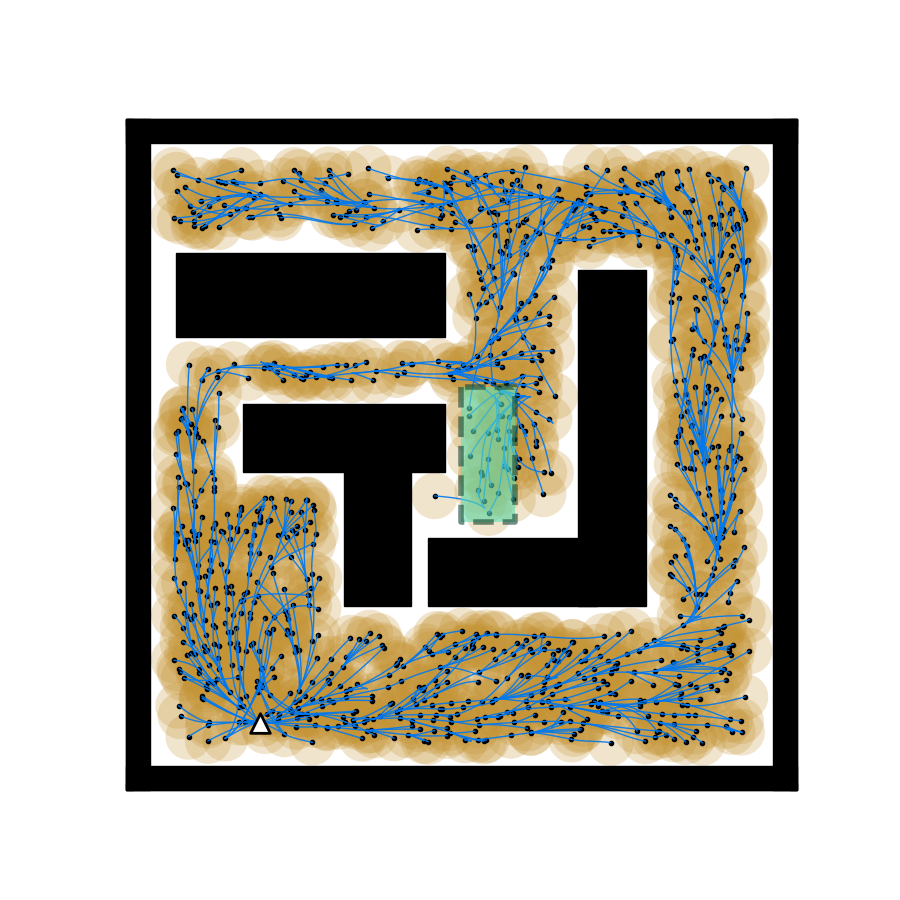}
        \caption{\centering \ransrrt{} tree with 1102 Nodes.}
        \label{fig:ransrrt_tree_1}
    \end{subfigure}%
    ~ 
    \begin{subfigure}[t]{0.48\linewidth}
        \centering
        \includegraphics[width=1.0\linewidth,trim={3cm 2cm 2cm 2cm},clip]{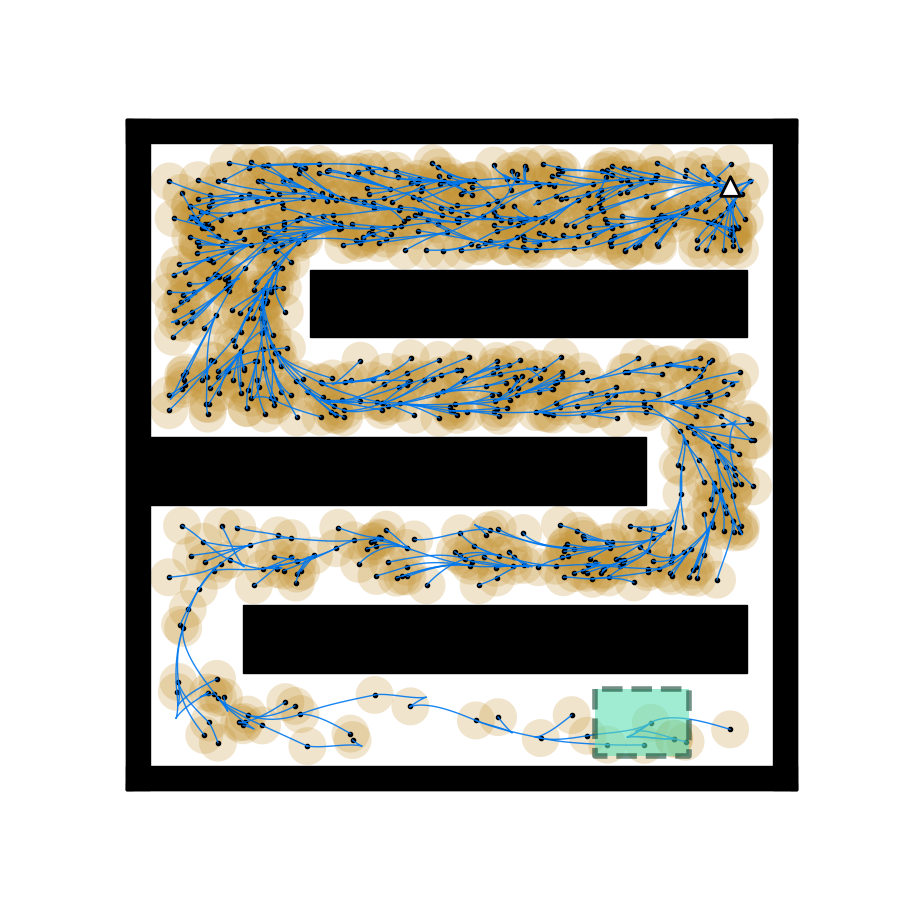}
        \caption{\centering \ransrrt{} tree with 802 Nodes.}
        \label{fig:ransrrt_tree_2}
    \end{subfigure}
    \begin{subfigure}[t]{0.48\linewidth}
        \centering
        \includegraphics[width=1.0\linewidth,trim={3cm 2cm 2cm 2cm},clip]{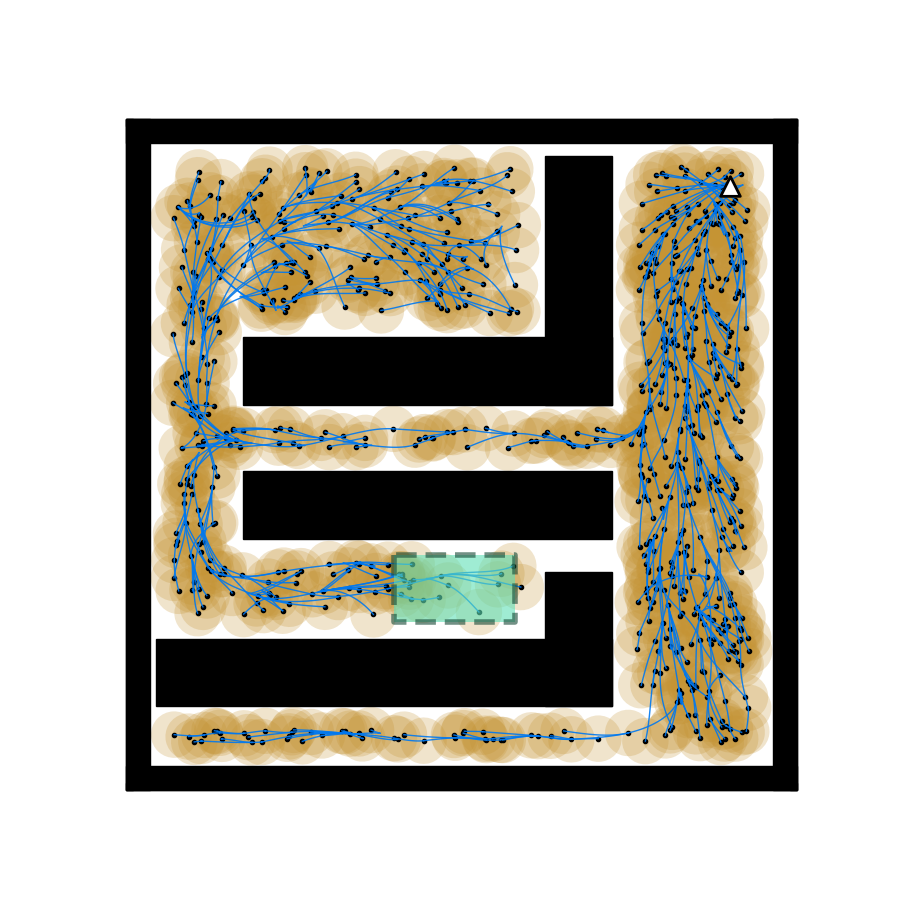}
        \caption{\centering \ransrrt{} tree with 865 Nodes.}
        \label{fig:ransrrt_tree_3}
    \end{subfigure}
    ~
    \begin{subfigure}[t]{0.48\linewidth}
        \centering
        \includegraphics[width=1.0\linewidth,trim={3cm 2cm 2cm 2cm},clip]{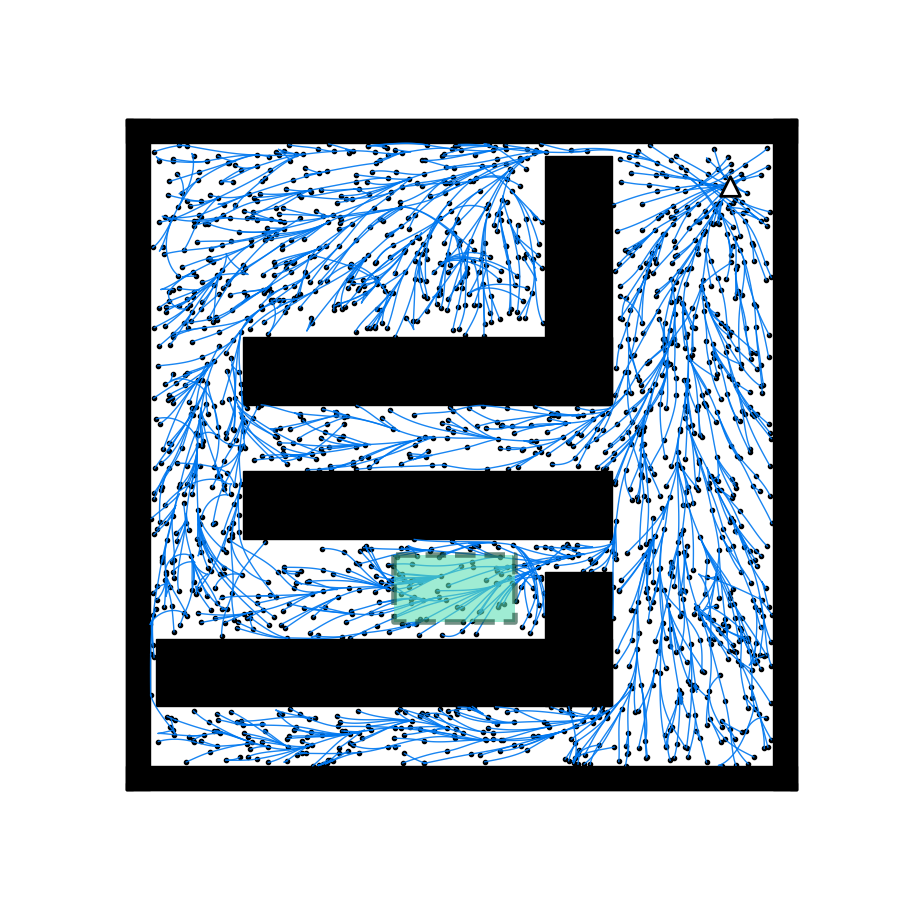}
        \caption{\centering \RRTs{} tree with 1791 Nodes.}
        \label{fig:rrt_tree}
    \end{subfigure}
    \caption{Trees grown by our proposed \ransrrt{} algorithm and standard \RRTs{} are demonstrated in Figures \ref{fig:ransrrt_tree_3} and \ref{fig:rrt_tree} respectively. The DR collision checks are represented by circles. The tree root is indicated by the white triangle and the goal region is the green rectangle with a dashed edge. Obstacles, including environment bounds are the black rectangles.}
\end{figure}

Constructing the \ransrrt{} trees in Figures \ref{fig:ransrrt_tree_1}-\ref{fig:ransrrt_tree_3} took 36, 23, and 28 minutes respectively. Each tree started with $2000$ randomly sampled nodes but only $1102$, $802$, and $865$ nodes were deemed feasible and safe and added to the trees, respectively. The \ransrrt{} trajectories were conservative in the sense that they avoided getting too close to obstacles. Small gaps, such as to the right of the goal in Figure \ref{fig:ransrrt_tree_3} or on the right side of the environment in Figure \ref{fig:ransrrt_tree_2}, were implicitly deemed too risky and avoided. On the other hand, a tree grown using the standard \RRTs{} in the same environment in Figure \ref{fig:rrt_tree} discovered such risky gaps and thereby returned an unsafe (risky) trajectory. In fact, it is easy to see that the optimal path to the goal in Figure \ref{fig:rrt_tree} scrapes by an obstacle and hence would result in a collisions even for small process noise.

\subsection{Low-Level Tracking Results}
\begin{figure}[h]
    \centering
    \begin{subfigure}[t]{0.49\linewidth}
        \centering
        \includegraphics[width=0.99\linewidth]{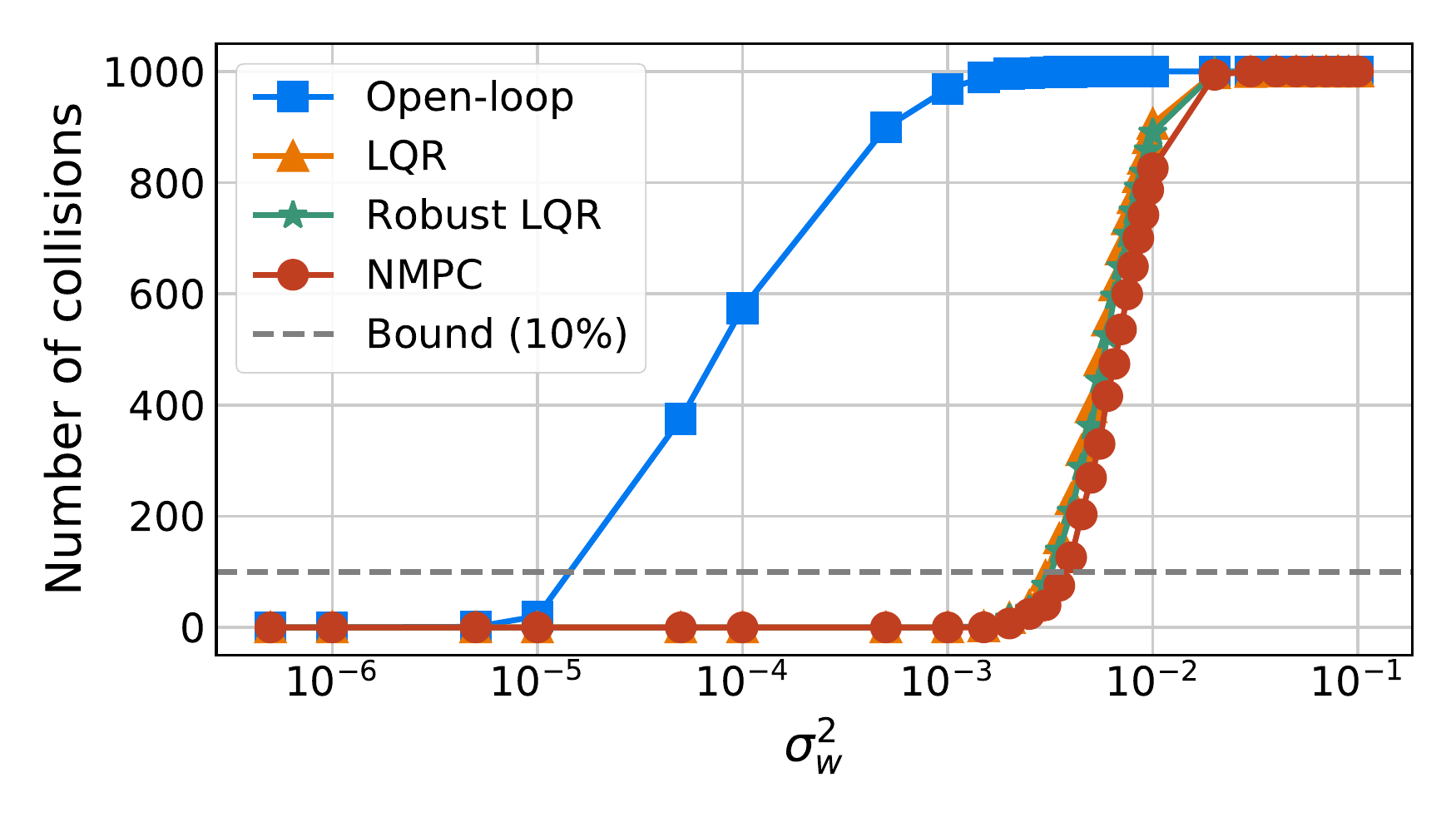}
        \caption{\centering Full view}
        \label{fig:monte_carlo_analysis_full}
    \end{subfigure}
    \begin{subfigure}[t]{0.49\linewidth}
        \centering
        \includegraphics[width=0.99\linewidth]{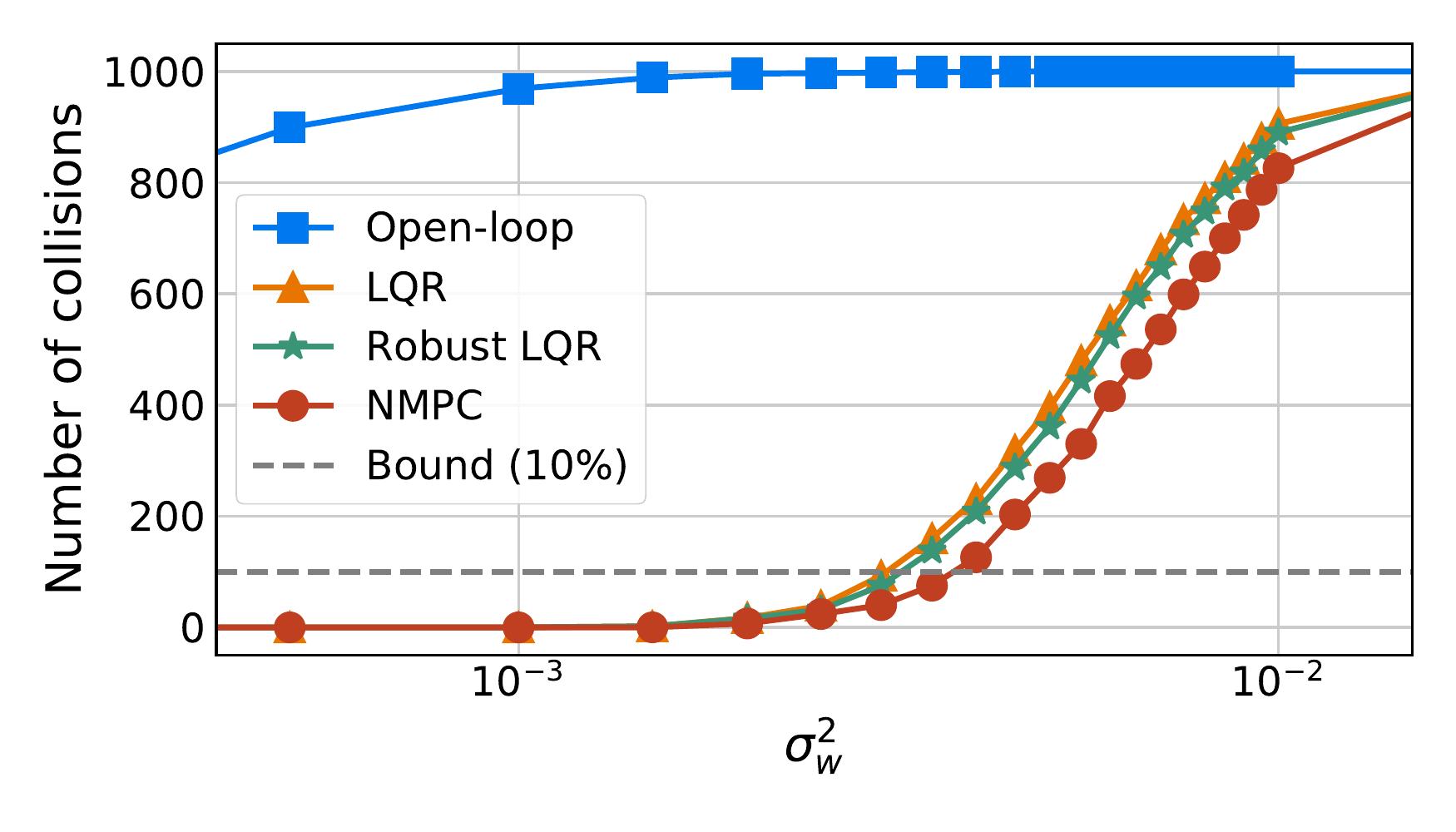}
        \caption{\centering Zoomed view (truncated x-axis).}
        \label{fig:monte_carlo_analysis_zoom}
    \end{subfigure}
    \caption{Number of failures for each controller across a range of noise covariance levels using distributionally robust collision checks. $1000$ Monte Carlo trials were run for each noise value. 
    }
    \label{fig:monte_carlo_analysis}
\end{figure}

We used open-loop control and three low-level closed-loop controllers 1) LQR, 2) robust LQR, and 3) NMPC, to track a high-level trajectory in the environment shown in Figure \ref{fig:ransrrt_tree_3} under realization of the Laplace noise. For each noise level, $1000$ Monte Carlo simulations were performed. The resulting trajectories are plotted in Figure \ref{fig:monte_carlo_analysis}. As expected, the noise level assumed for the high-level plan $\sigma_w^2 = 5\cdot 10^{-7}$ was insignificant: even open-loop control succeeded.
However, as the noise level increased, open-loop control began to fail. At around $\sigma^2_w = 0.001$, the open-loop control almost always failed, while the other controllers almost always succeeded. From there, the feedback controllers began failing more frequently. Robust LQR did slightly better than LQR with fewer collisions for each noise level. Both were outperformed by NMPC which was better able to reject the more aggressive noise, leading to notably fewer collisions. 
The largest noise levels tested for which the plan failure risk bound of $10$ percent was satisfied were
$0.003$ for LQR and robust LQR and
$0.0035$ for NMPC. These are, respectively, $6000$ and $7000$ folds larger than the assumed noise covariance $\sigma^2_w = 0.0000005$.

\begin{figure}[h]
    \centering
    \begin{subfigure}[t]{0.49\linewidth}
        \centering
        \includegraphics[width=0.99\linewidth]{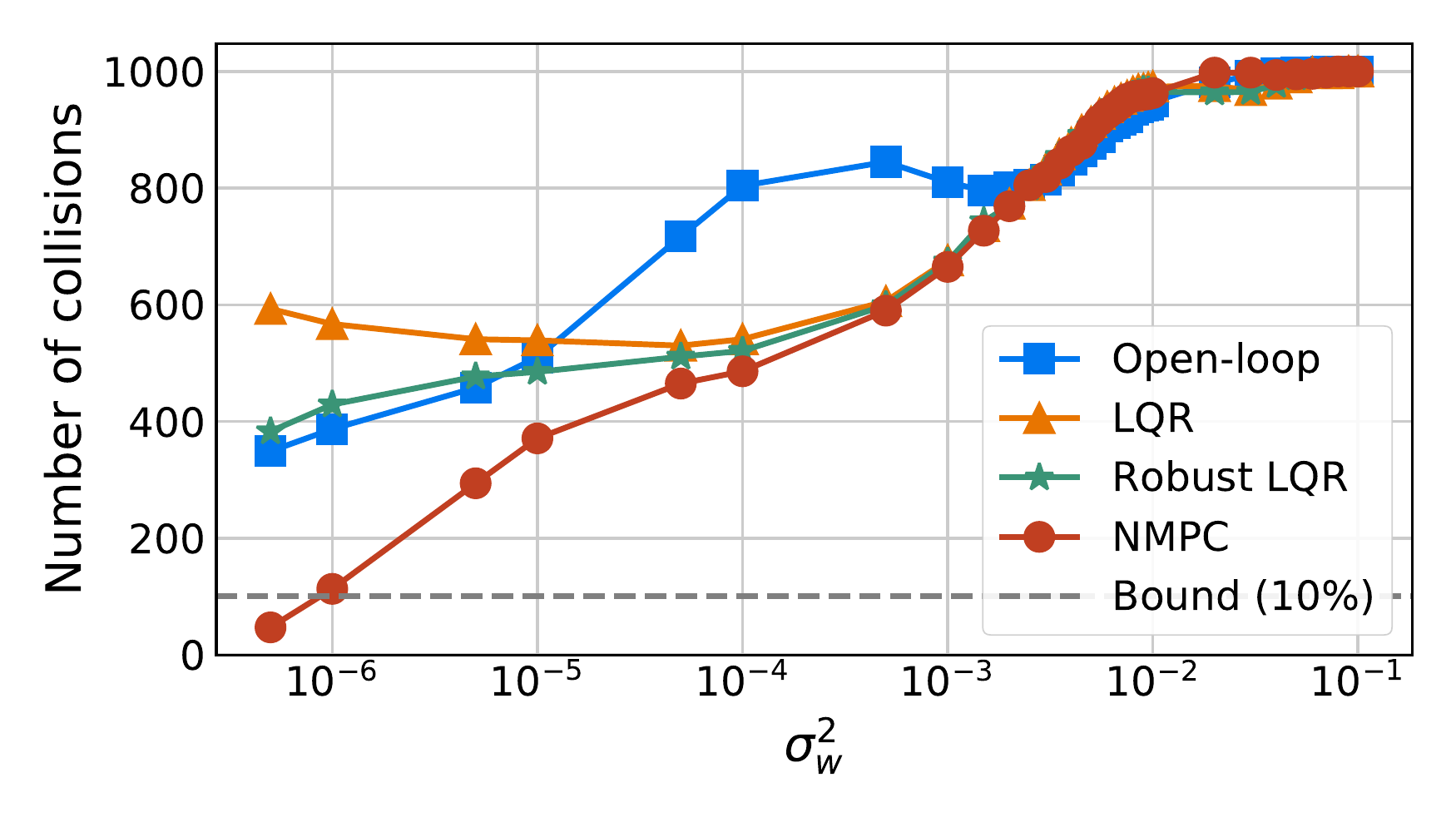}
        \caption{\centering Full view}
        \label{fig:monte_carlo_analysis_full_NoDR}
    \end{subfigure}
    \begin{subfigure}[t]{0.49\linewidth}
        \centering
        \includegraphics[width=0.99\linewidth]{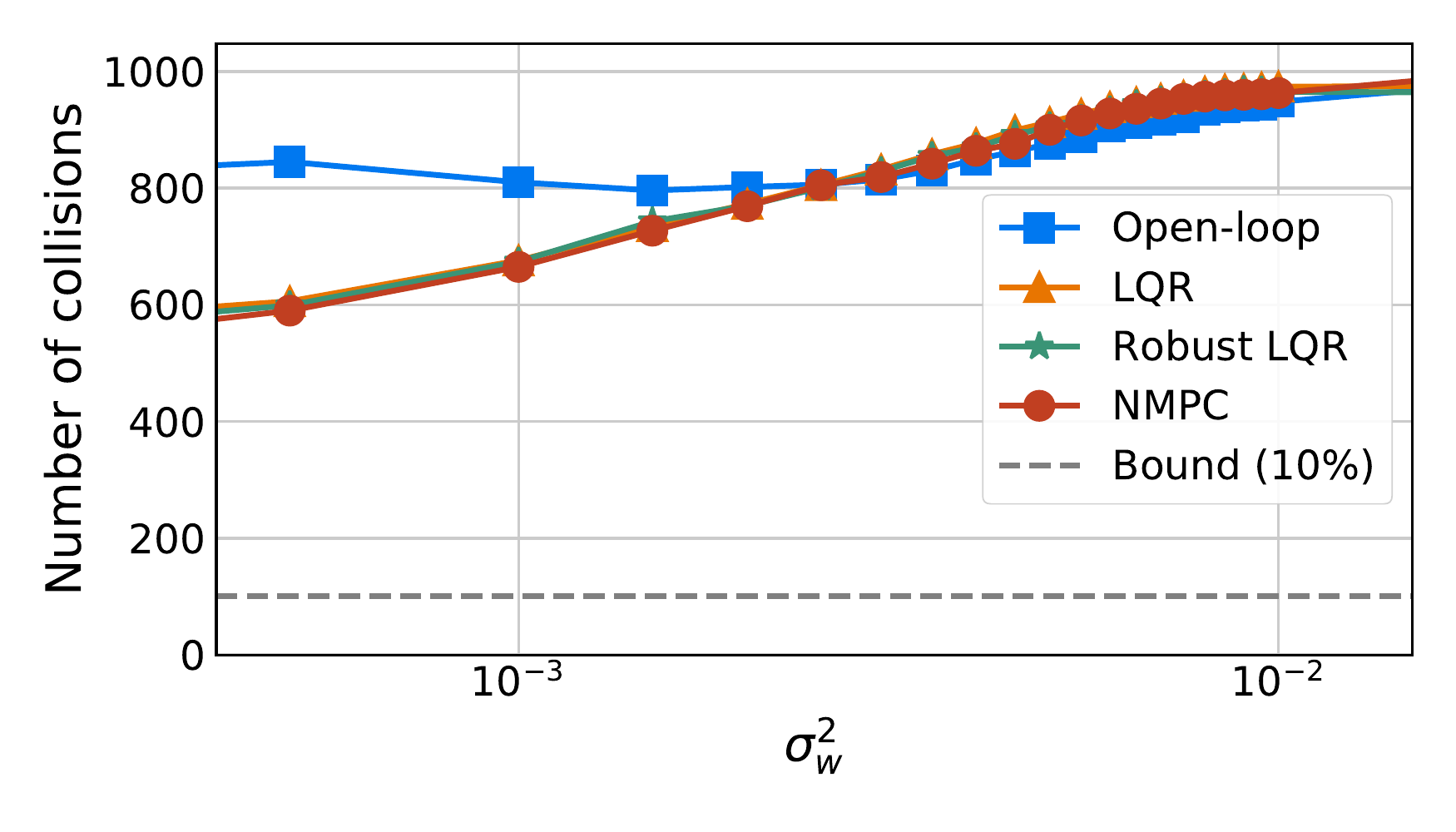}
        \caption{\centering Zoomed view (truncated x-axis).}
        \label{fig:monte_carlo_analysis_zoom_NoDR}
    \end{subfigure}
    \caption{Number of failures for each controller across a range of noise covariance levels without using distributionally robust obstacle padding. $1000$ Monte Carlo trials were run for each noise value.
    }
    \label{fig:monte_carlo_analysis_NoDR}
\end{figure}

\begin{figure}[p!]
    \centering
    \begin{subfigure}[t]{0.48\linewidth}
        \centering
        \includegraphics[width=0.99\linewidth]{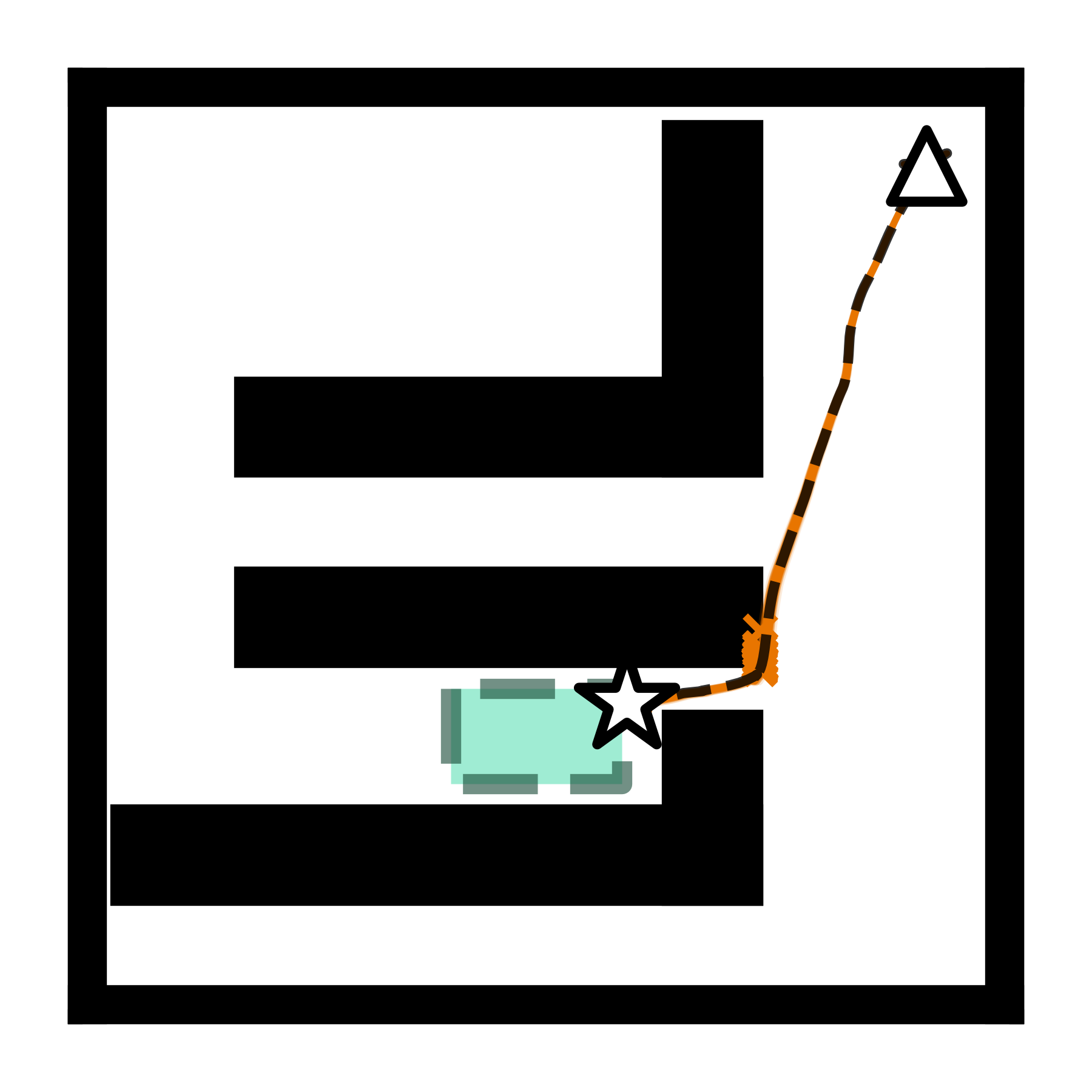}
        \caption{\centering Open-loop}
        \label{fig:path_plot_open_loop_0p0000005_NoDR}
    \end{subfigure}%
    ~
    \begin{subfigure}[t]{0.48\linewidth}
        \centering
        \includegraphics[width=0.99\linewidth]{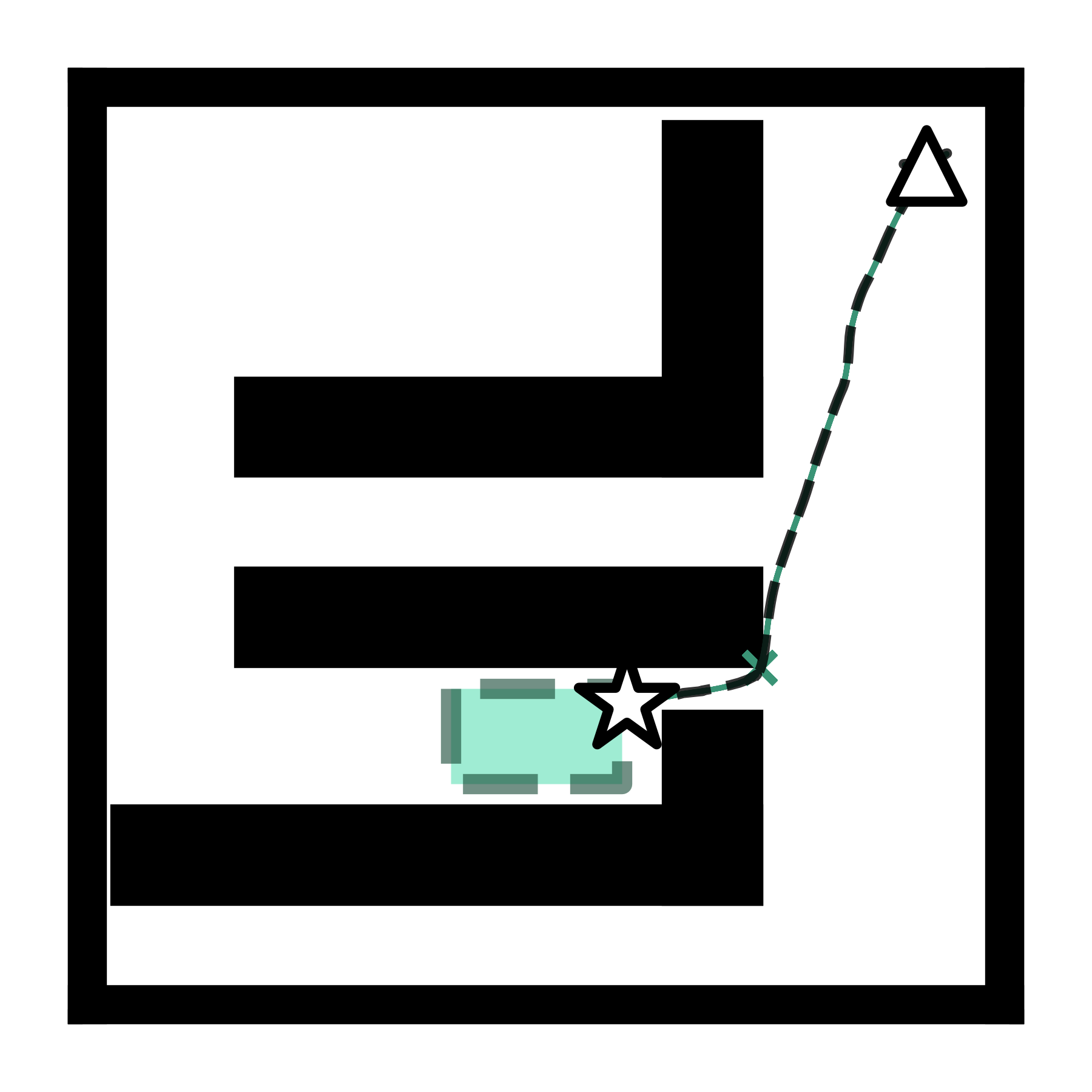}
        \caption{\centering LQR}
        \label{fig:path_plot_lqr_0p0000005_NoDR}
    \end{subfigure}
    ~
    \begin{subfigure}[t]{0.48\linewidth}
        \centering
        \includegraphics[width=0.99\linewidth]{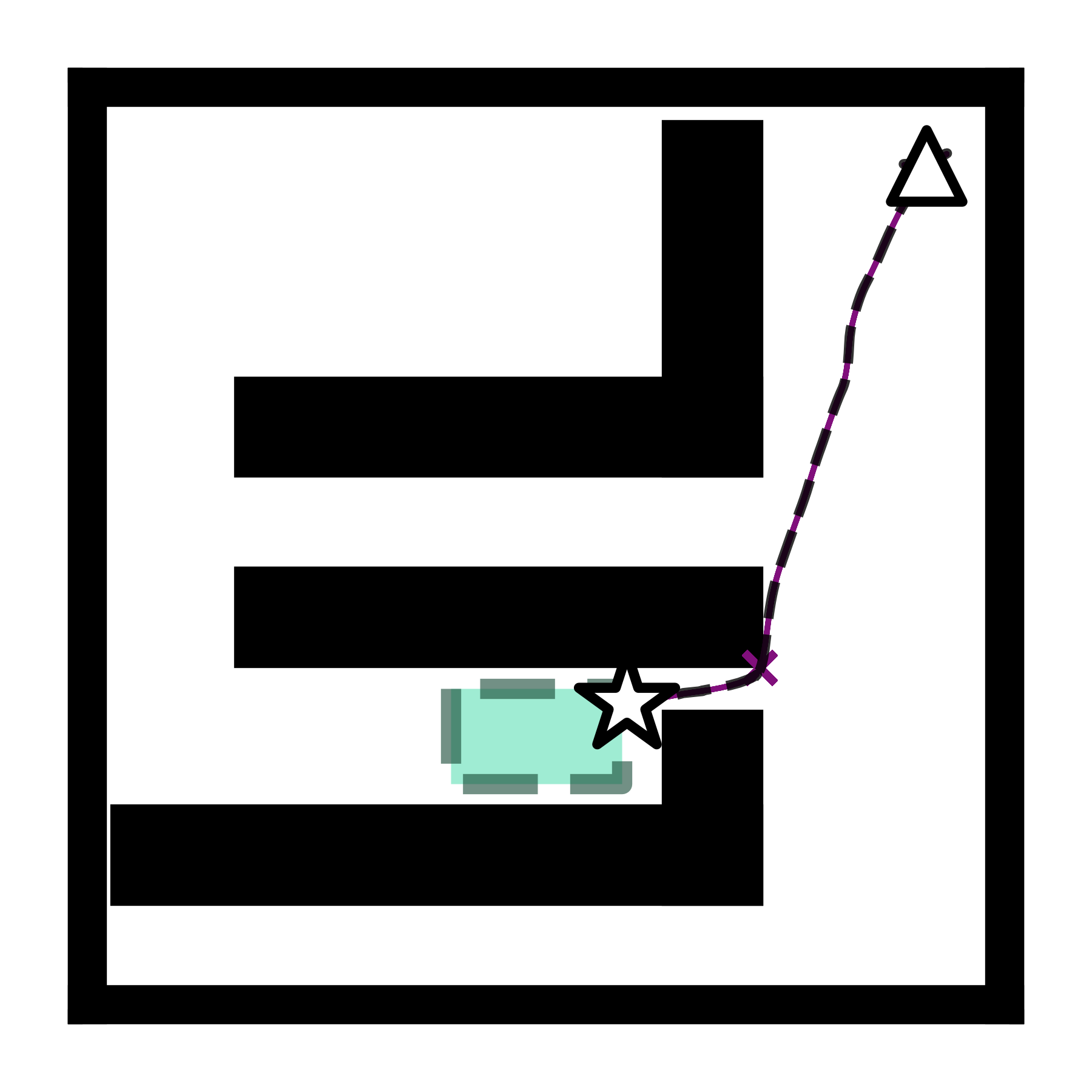}
        \caption{\centering LQRm}
        \label{fig:path_plot_lqrm_0p0000005_NoDR}
    \end{subfigure}%
    ~
    \begin{subfigure}[t]{0.48\linewidth}
        \centering
        \includegraphics[width=0.99\linewidth]{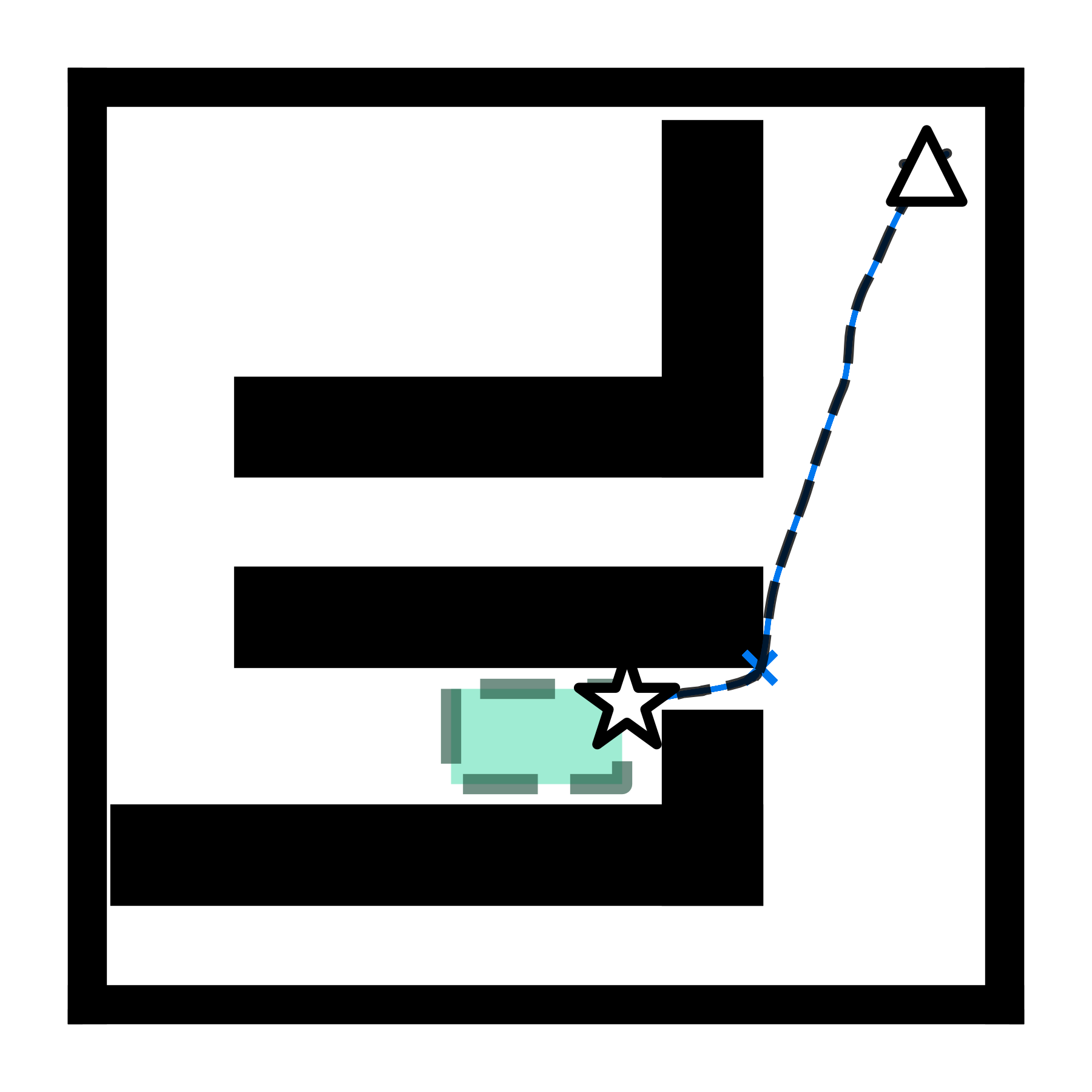}
        \caption{\centering NMPC}
        \label{fig:path_plot_nmpc_0p0000005_NoDR}
    \end{subfigure}
    \caption{The results of 1000 independent Monte Carlo trials with different low-level reference tracking controllers. The reference trajectory was planned \textbf{without} distributionally robust obstacle padding, and is shown as a dashed line. The start and goal locations are marked by triangle and star icons respectively. The goal area is a green rectangular area whose border is marked by a dashed line. The terminal state of failed and successful trajectories are shown by `O' and `X' markers respectively. Obstacles and the free space are represented by solid black and white regions, respectively. The disturbance variance was $\sigma_w^2 = 0.0000005$.}
    \label{fig:path_plot_0p0000005_NoDR}
\end{figure}

\begin{figure}[p!]
    \centering
    \begin{subfigure}[t]{0.48\linewidth}
        \centering
        \includegraphics[width=0.99\linewidth]{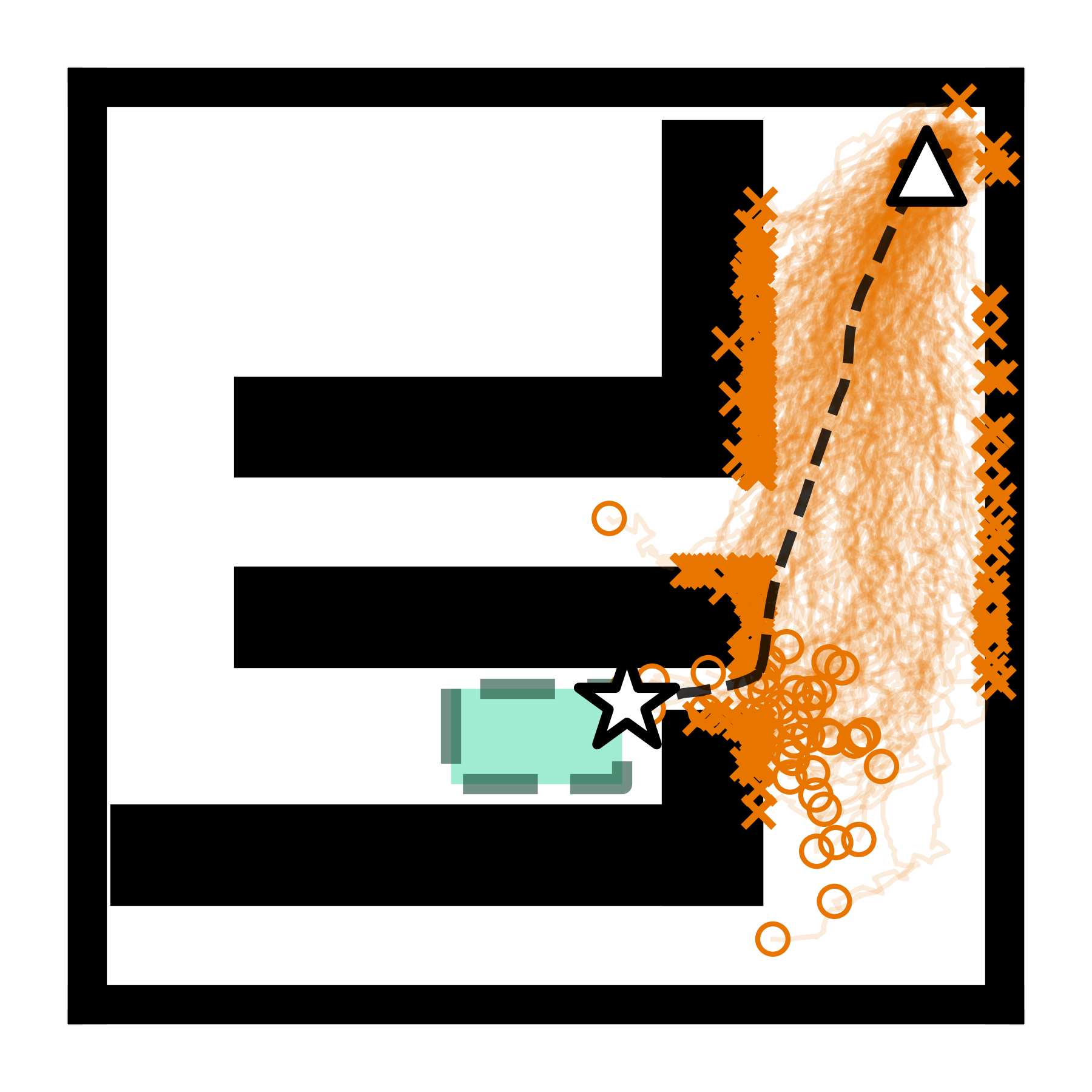}
        \caption{\centering Open-loop}
        \label{fig:path_plot_open_loop_0p003_NoDR}
    \end{subfigure}%
    ~
    \begin{subfigure}[t]{0.48\linewidth}
        \centering
        \includegraphics[width=0.99\linewidth]{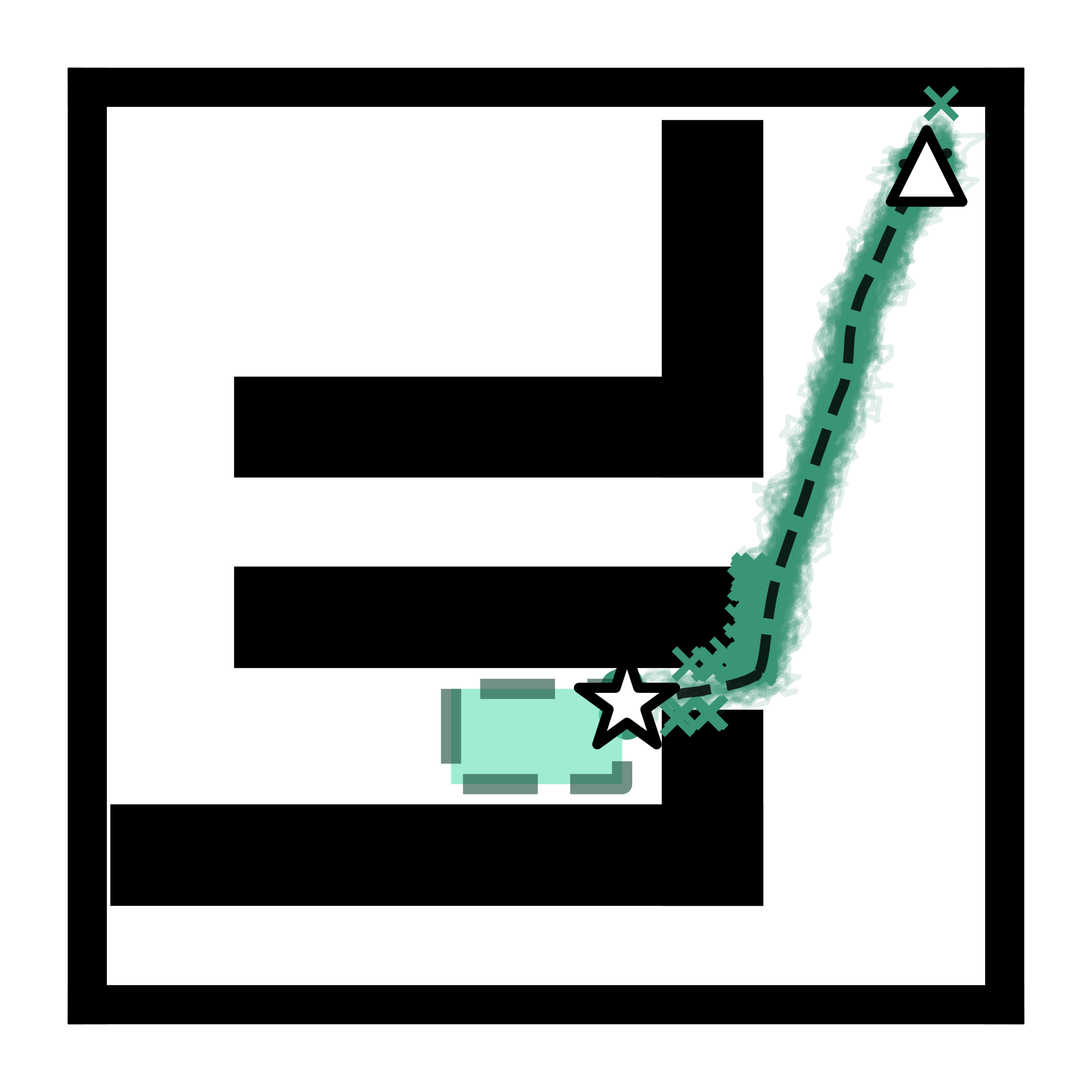}
        \caption{\centering LQR}
        \label{fig:path_plot_lqr_0p003_NoDR}
    \end{subfigure}
    ~
    \begin{subfigure}[t]{0.48\linewidth}
        \centering
        \includegraphics[width=0.99\linewidth]{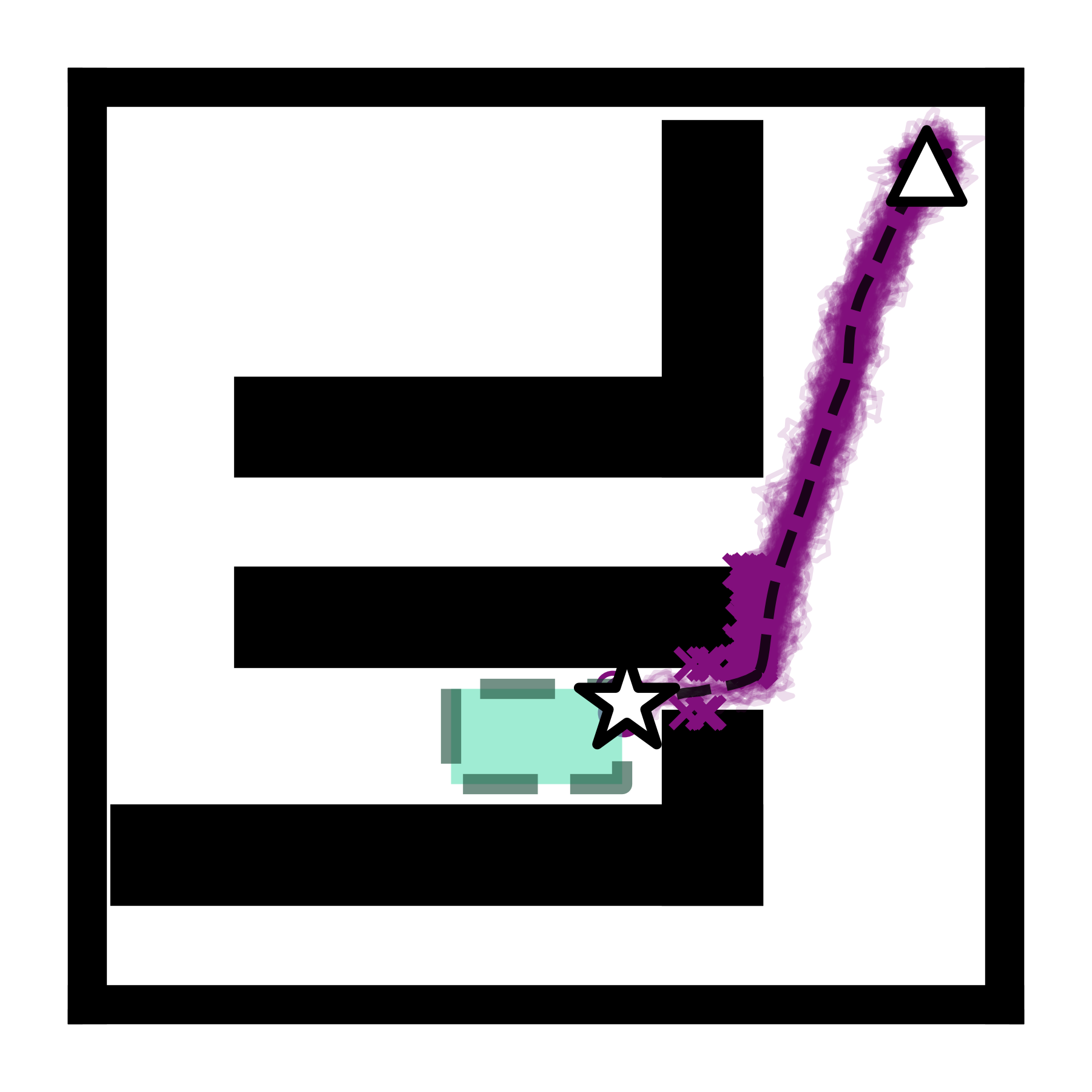}
        \caption{\centering LQRm}
        \label{fig:path_plot_lqrm_0p003_NoDR}
    \end{subfigure}
    ~
    \begin{subfigure}[t]{0.48\linewidth}
        \centering
        \includegraphics[width=0.99\linewidth]{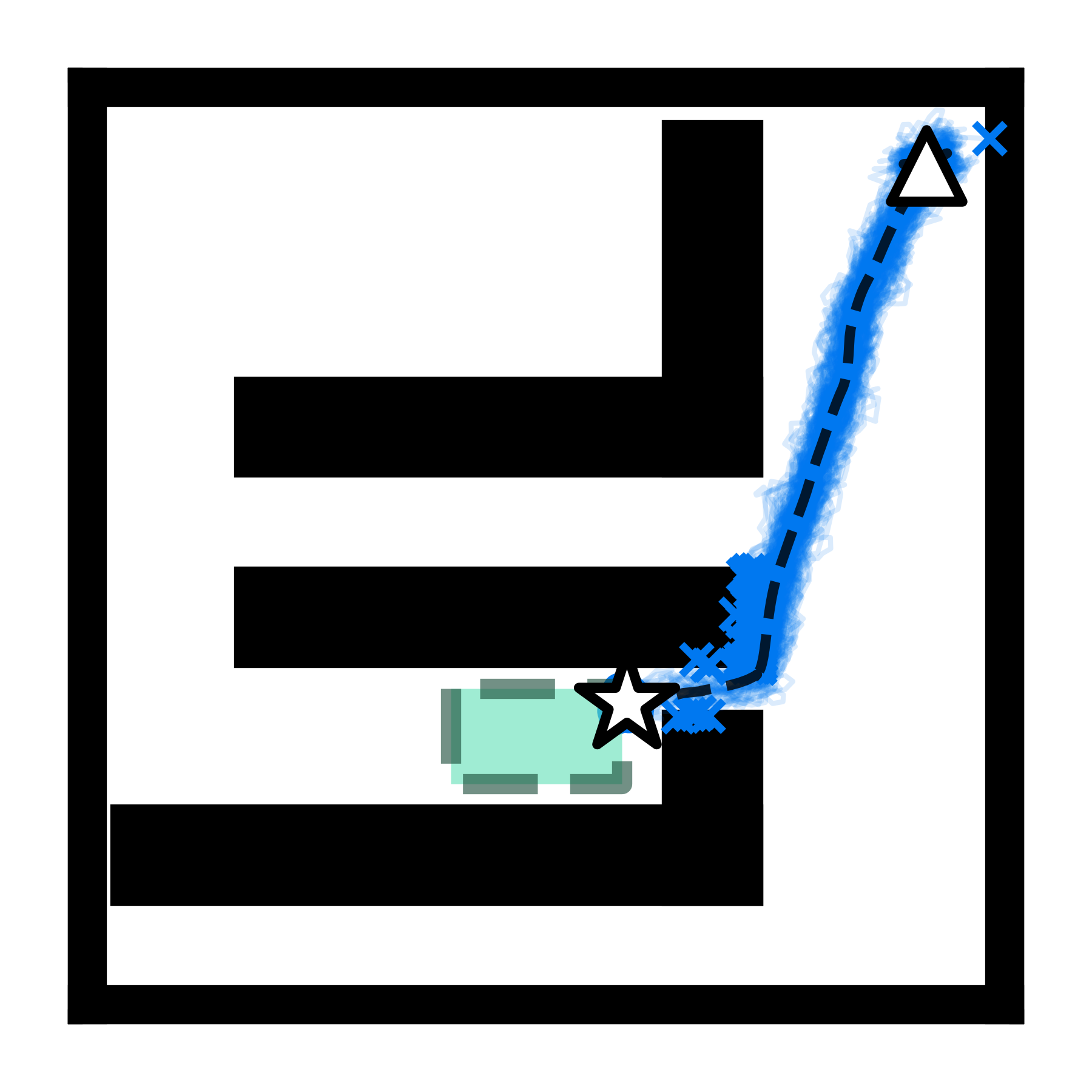}
        \caption{\centering NMPC}
        \label{fig:path_plot_nmpc_0p003_NoDR}
    \end{subfigure}
    \caption{The results of 1000 independent Monte Carlo trials with different low-level reference tracking controllers. The reference trajectory was planned \textbf{without} distributionally robust obstacle padding, and is shown as a dashed line. The start and goal locations are marked by triangle and star icons respectively. The goal area is a green rectangular area whose border is marked by a dashed line. The terminal state of failed and successful trajectories are shown by `O' and `X' markers respectively. Obstacles and the free space are represented by solid black and white regions, respectively. The disturbance variance was $\sigma_w^2 = 0.003$.}    
    \label{fig:path_plot_0p003_NoDR}
\end{figure}

For comparison, we ran the same experiment, but with the distributionally robust obstacle padding disabled. The planner returned trajectories which passed near obstacle boundaries, as shown in the tree in Figure \ref{fig:rrt_tree}, with realized trajectories plotted in Figures \ref{fig:path_plot_0p003_NoDR} and \ref{fig:path_plot_0p0000005_NoDR}. Consequently, the reference trajectory itself was extremely unsafe, as evidenced by the high failure rates exhibited in Figure \ref{fig:monte_carlo_analysis_NoDR}. Only at an extremely low noise level $\sigma^2_w = 0.0000005$ was NMPC able to reliably avoid collisions, while the LQR and LQRm controllers failed to do so. However with slightly more noise at the level $\sigma^2_w = 0.00001$, all control schemes led to significant probability of collision (greater than $35\%$). By contrast, the \ransrrt{} reference trajectory was so safe that even open-loop control led to a low probability of collision (less than $10\%$) at this noise level $\sigma^2_w = 0.00001$. This trend continued all the way up through the noise level $\sigma^2_w = 0.001$ with the probability of collision along the non-robust reference trajectory continuing to degrade, while the robust reference trajectory remained nearly perfectly safe with any of the low-level feedback controllers, thus demonstrating the clear benefit of \ransrrt{} over vanilla \RRTs{}. Eventually the noise became too powerful and collisions became a near certainty regardless of the planner or controller used.

\begin{figure}[p!]
    \centering
    \begin{subfigure}[t]{0.48\linewidth}
        \centering
        \includegraphics[width=0.99\linewidth]{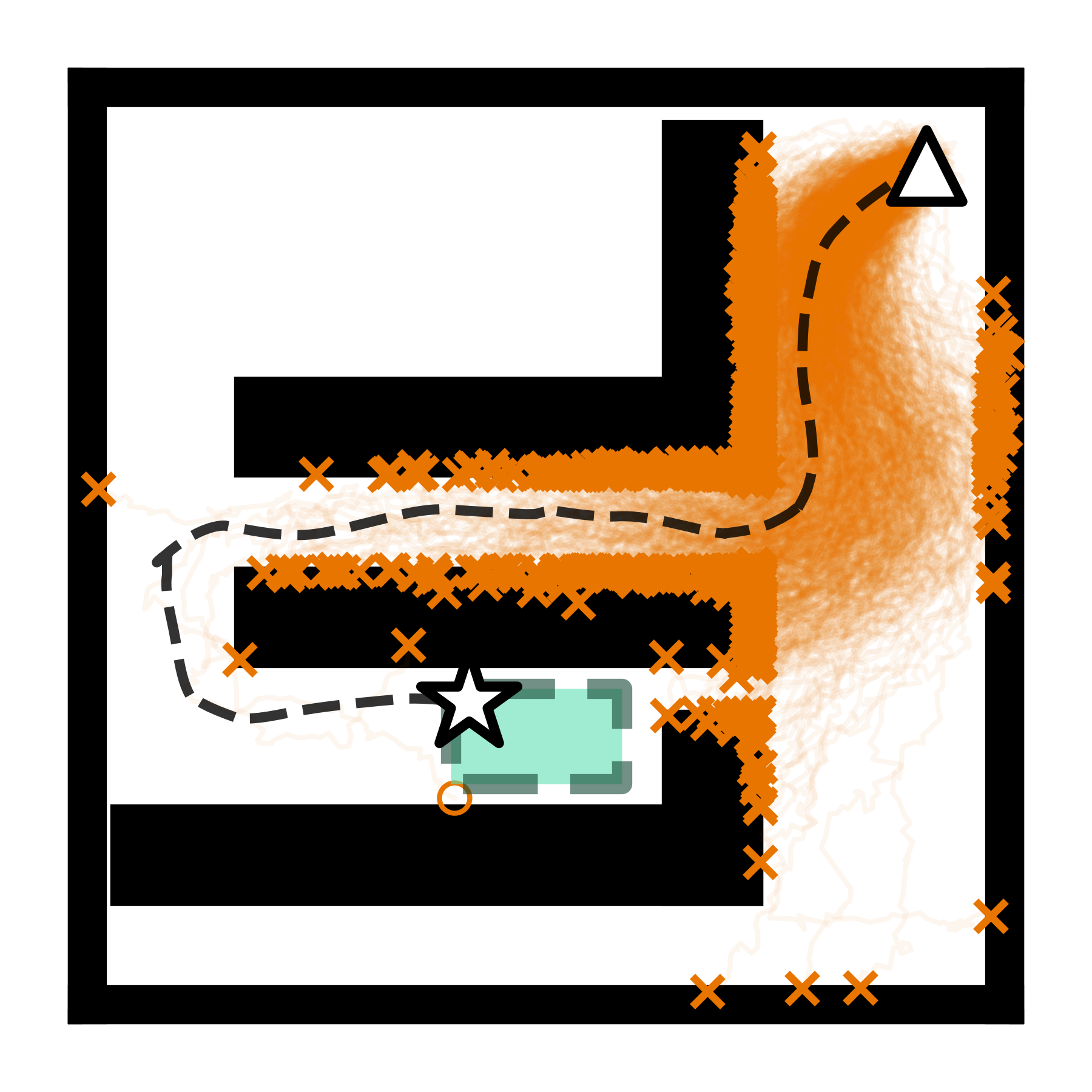}
        \caption{\centering Open-loop}
        \label{fig:path_plot_open_loop_0p0035}
    \end{subfigure}%
    ~
    \begin{subfigure}[t]{0.48\linewidth}
        \centering
        \includegraphics[width=0.99\linewidth]{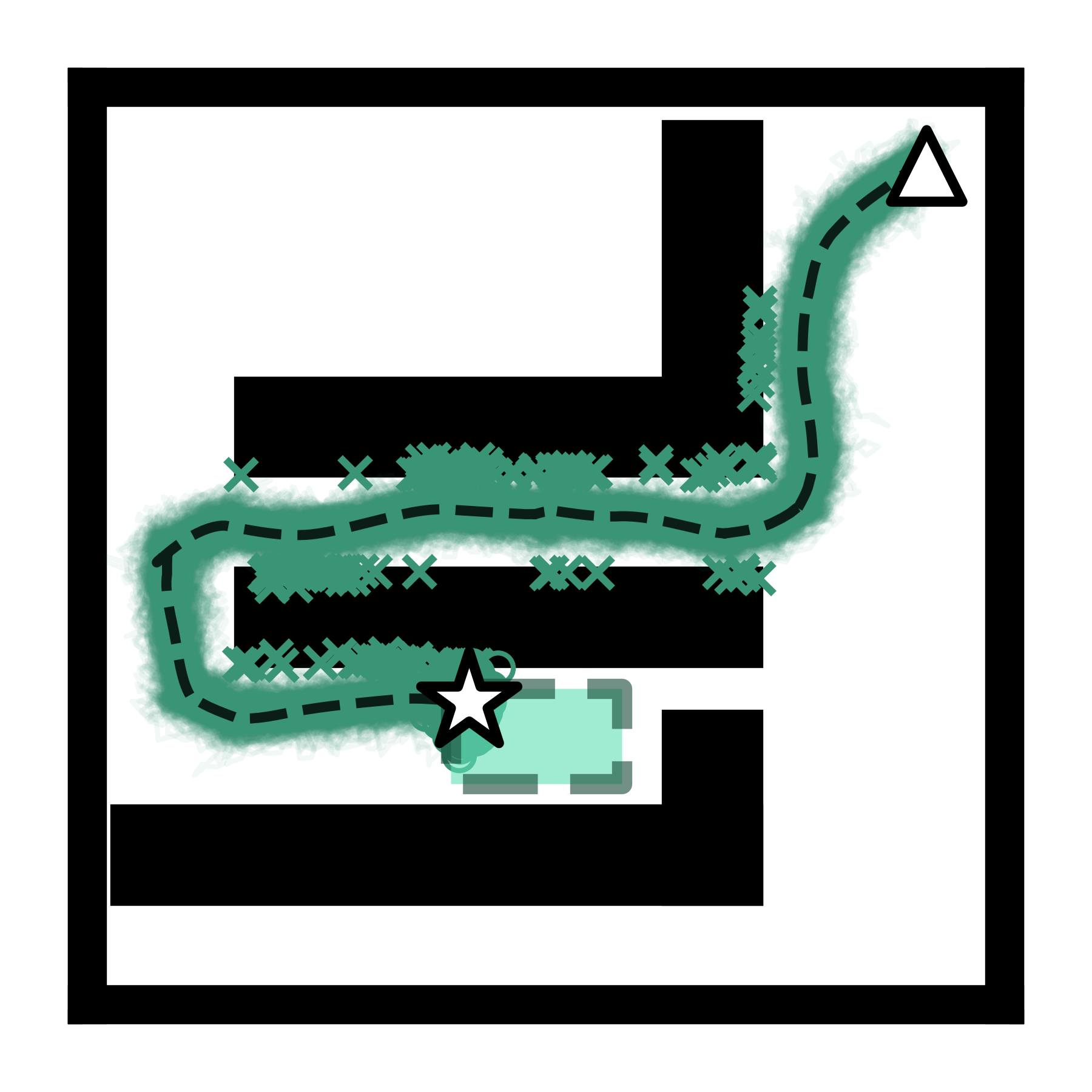}
        \caption{\centering LQR}
        \label{fig:path_plot_lqr_0p0035}
    \end{subfigure}
    ~
    \begin{subfigure}[t]{0.48\linewidth}
        \centering
        \includegraphics[width=0.99\linewidth]{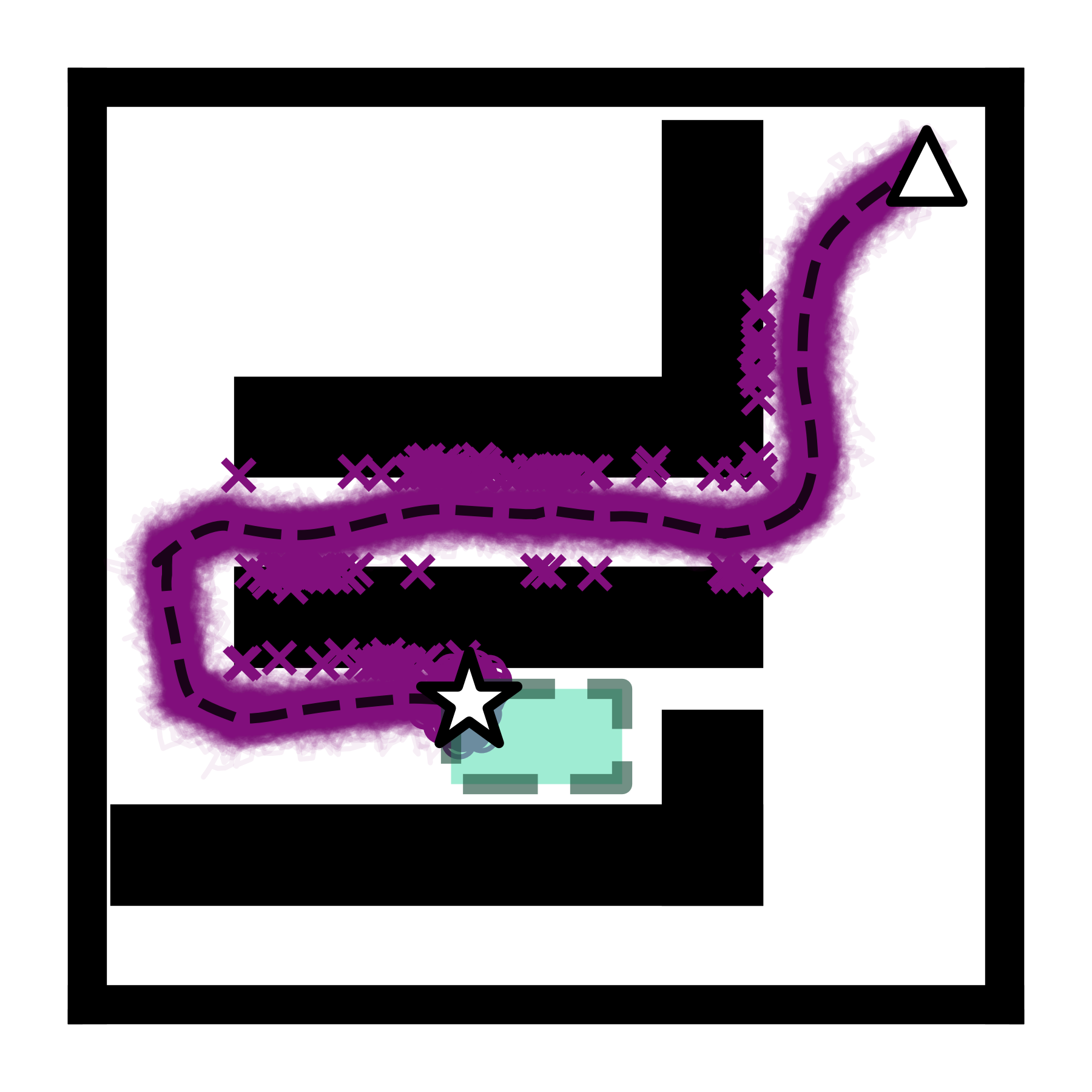}
        \caption{\centering LQRm}
        \label{fig:path_plot_lqrm_0p0035}
    \end{subfigure}%
    ~
    \begin{subfigure}[t]{0.48\linewidth}
        \centering
        \includegraphics[width=0.99\linewidth]{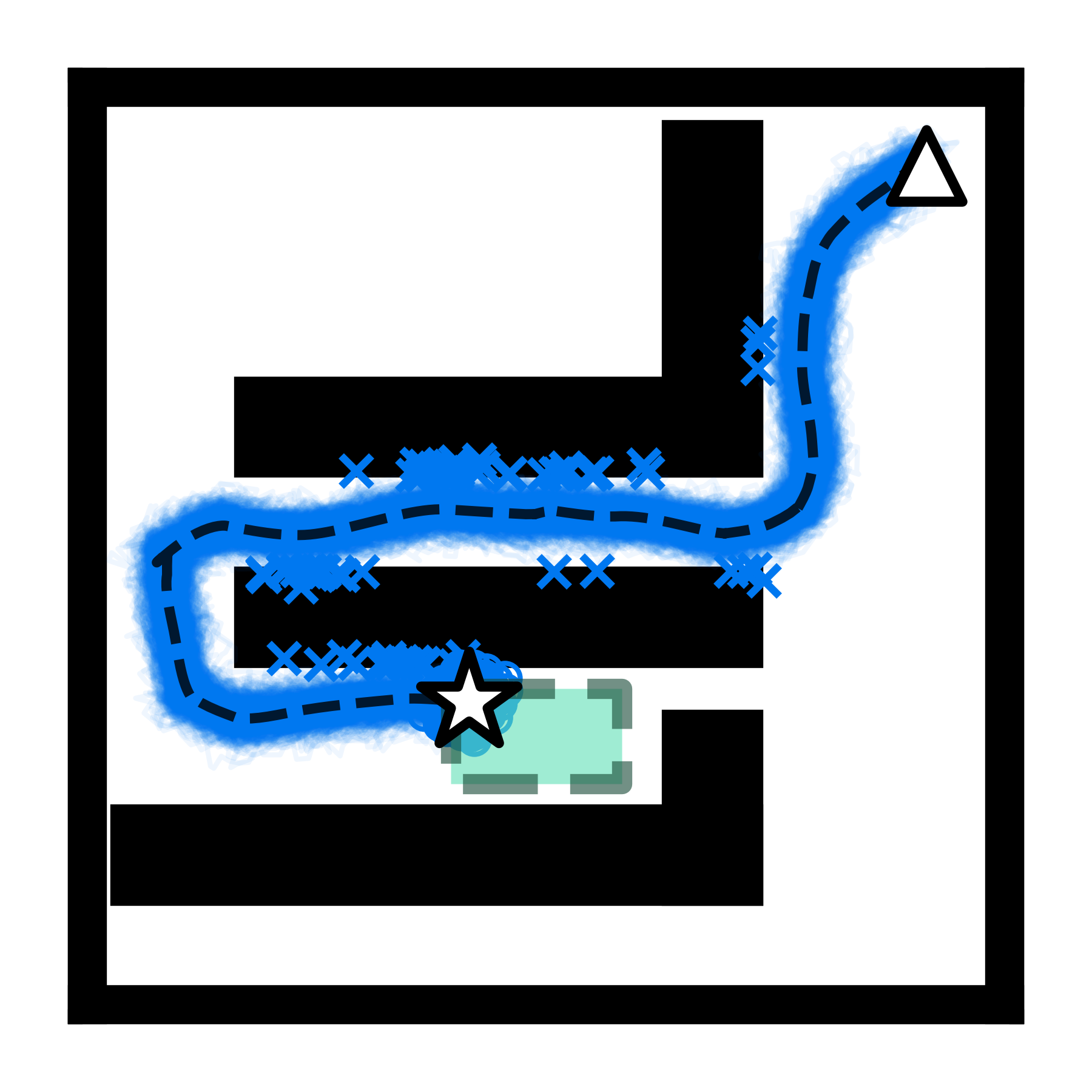}
        \caption{\centering NMPC}
        \label{fig:path_plot_nmpc_0p0035}
    \end{subfigure}
    \caption{The results of 1000 independent Monte Carlo trials with different low-level reference tracking controllers. The reference trajectory was planned \textbf{with} distributionally robust obstacle padding, and is shown as a dashed line. The start and goal locations are marked by triangle and star icons respectively. The goal area is a green rectangular area whose border is marked by a dashed line. The terminal state of failed and successful trajectories are shown by `O' and `X' markers respectively. Obstacles and the free space are represented by solid black and white regions, respectively. The disturbance variance was $\sigma_w^2 = 0.0035$.}    
    \label{fig:path_plot_0p0035}
\end{figure}

In Figure \ref{fig:path_plot_0p0035} the realized trajectories obtained through the Monte Carlo simulations are plotted for $\sigma_w^2 = 0.0035$. The performance metrics for $\sigma_w^2 = 0.0000005$ and $\sigma_w^2 = 0.0035$ are tabulated in Tables \ref{tab:path_table_0p0000005} and \ref{tab:path_table_0p0035} respectively. We use the following metrics to evaluate the effectiveness of each tracking controller:
\begin{enumerate}
    \item The number of collisions:  the number of Monte Carlo trials in which the realized trajectory collided with an obstacle.
    \item The average $\delta_x$ and $u$ costs: the realized state-deviation cost $\sum_{k=0}^{T} \delta_x[k]^\tp Q[k] \delta_x[k]$ and control cost $\sum_{k=0}^{T-1} u[k]^\tp R u[k]$, conditionally averaged across all collision-free trajectories.
    \item The average run time: the time taken to run the entire Monte Carlo trial, conditionally averaged across all collision-free trajectories, which is necessary as simulations terminate immediately upon collision. 
\end{enumerate}
With $\sigma_w^2 = 0.0000005$, the trajectory was short enough so that collisions did not occur even with purely open-loop control, although the trajectories began to diverge from the reference. The closed-loop controllers exhibited minimal state deviations, as reflected in the significantly lower average $\delta_x$ cost. The difference in $\delta_x$ and $u$ costs between each closed-loop controller was insignificant; since the state remained extremely close to the reference, the inputs generated by each controller were very similar.

However with $\sigma_w^2 = 0.0035$ as shown in Figure \ref{fig:path_plot_0p0035}, the following observations were made. 
Almost all open-loop trajectories ended with collisions as shown in Figure \ref{fig:path_plot_open_loop_0p0035}. Compared to the standard LQR, robust LQR led to a lower number of collisions and state-deviation cost as shown in Figures \ref{fig:path_plot_lqr_0p0035}, \ref{fig:path_plot_lqrm_0p0035} but both were significantly outperformed by NMPC. NMPC trajectories generally remained closer to the reference than LQR or robust LQR along with lower number of collision as the robot moved through the corridor as shown in Figure \ref{fig:path_plot_nmpc_0p0035}. However, NMPC's closer reference tracking and collision-avoidance came at a price, as it used more control effort than LQR and robust LQR. Likewise, the more sophisticated computations involved in solving the NLPs in NMPC led to a longer average run time. It is evident that under both the noise settings, the NMPC outperforms other controllers in tracking the given reference trajectory to reach the goal with low failure rate and a better cost. 

\begin{table}[h]
    \centering
    \begin{tabular}{| c | c | c | c | c |} 
    \hline
        Controller & Num. collisions & $\delta_x$ cost & $u$ cost & Run time (s) \\
        \hline \hline
        Open-loop & 0 & 30.869  & 54.989  & 0.0103  \\
        LQR & 0 & 3.403  & 43.267   & 0.1867  \\
        LQRm & 0 & 2.911  & 44.351  & 0.2833 \\
        NMPC & 0 & 3.522 & 43.240 & 5.6234 \\
        \hline
    \end{tabular}
    \caption{Performance metrics for all controllers from Monte Carlo simulation with $\sigma_w^2 = 0.0000005$.}
    \label{tab:path_table_0p0000005}
\end{table}

\begin{table}[h]
    \centering
    \begin{tabular}{| c | c | c | c | c |} 
    \hline
        Controller & Num. collisions & $\delta_x$ cost & $u$ cost & Run time (s) \\
        \hline \hline
        Open-loop & 999 & 5103.148  & 54.989  & 0.0018 \\
        LQR & 160 & 827.949  & 110.016  & 0.1751 \\
        LQRm & 138 & 820.916 & 114.646   & 0.2946 \\
        NMPC & 75 & 732.646 & 131.059  & 6.9159 \\
        \hline
    \end{tabular}
    \caption{Performance metrics for all controllers from Monte Carlo simulation with $\sigma_w^2 = 0.0035$.}
    \label{tab:path_table_0p0035}
\end{table}

\section{Conclusion and Future Work}\label{sec:conc}
We proposed a risk-averse control architecture tailored for safely controlling stochastic nonlinear robotic systems, which combines
a novel nonlinear steering-based variant of \RRTs{} called \ransrrt{} that accounts for risk by performing DR collision checks
with low-level reference tracking controllers. 
We performed thorough numerical experiments using unicycle dynamics, compared three controllers, and observed better performance from NMPC than LQR variants.
We showed that the despite the usage of very small noise level assumptions in the high-level planner, the low-level controllers performed well under moderate and aggressive disturbance realizations. Future research involves considering a full nonlinear sensor model while incorporating the exact DR risk constraints in the optimization problems of both the levels of autonomy stack for accurate risk assessment. We will also seek to decrease the computation time of NMPC through the usage of code generation tools.

\bibliographystyle{IEEEtranS}

\bibliography{references}

\clearpage

\clearpage
\begin{figure}[p!]
    \centering
    \begin{subfigure}[t]{0.48\linewidth}
        \centering
        \includegraphics[width=0.99\linewidth]{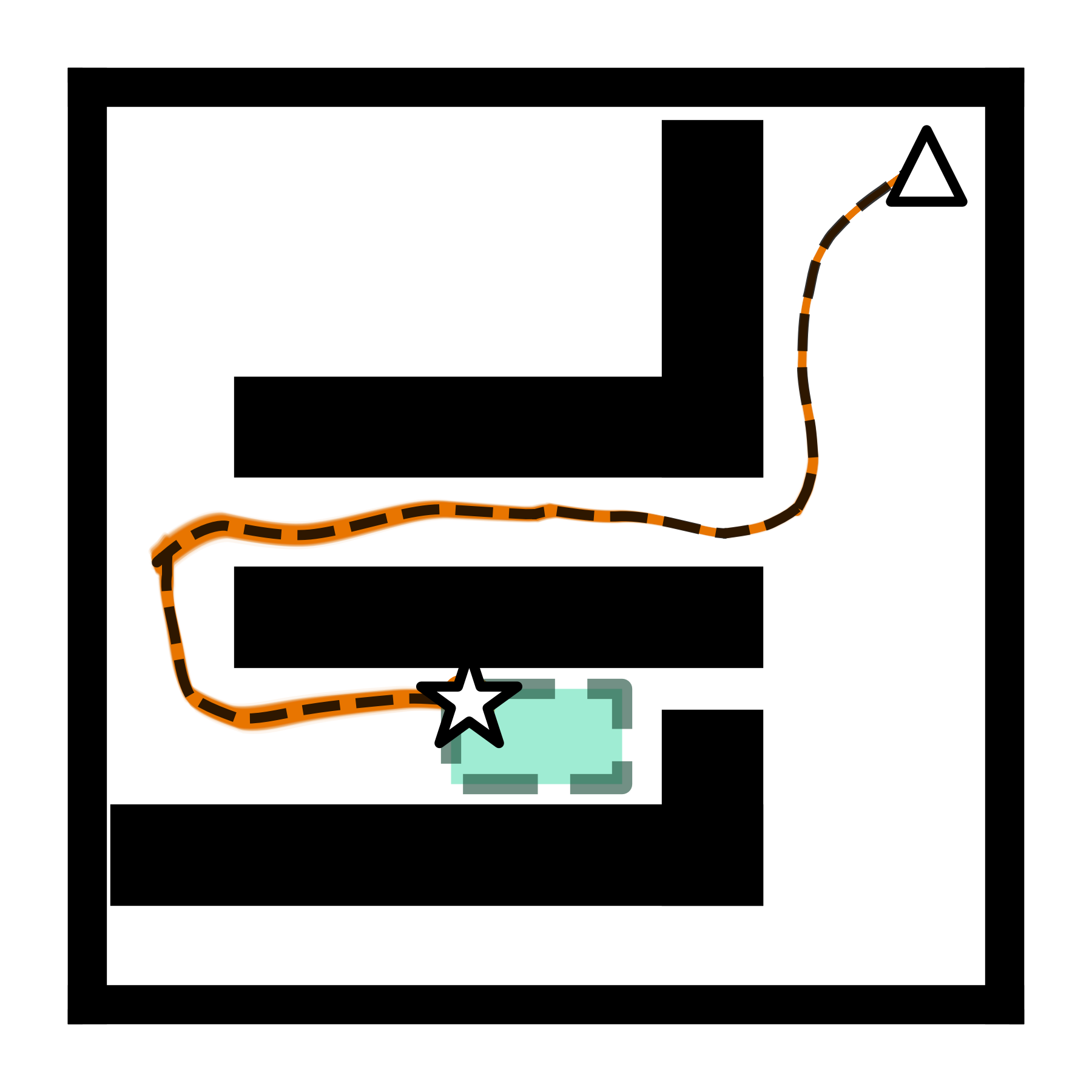}
        \caption{\centering Open-loop}
        \label{fig:path_plot_open_loop_0p0000005}
    \end{subfigure}%
    ~
    \begin{subfigure}[t]{0.48\linewidth}
        \centering
        \includegraphics[width=0.99\linewidth]{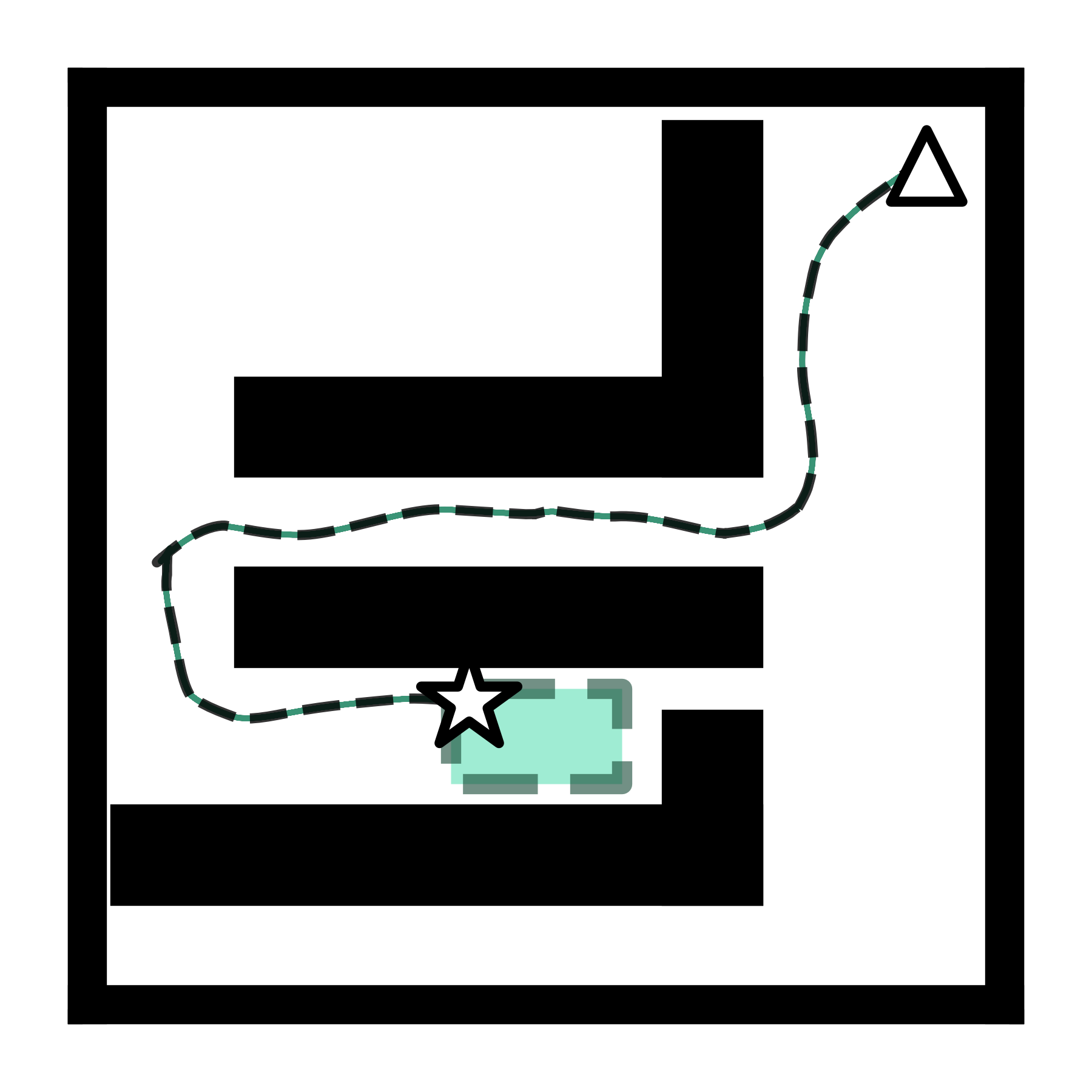}
        \caption{\centering LQR}
        \label{fig:path_plot_lqr_0p0000005}
    \end{subfigure}
    ~
    \begin{subfigure}[t]{0.48\linewidth}
        \centering
        \includegraphics[width=0.99\linewidth]{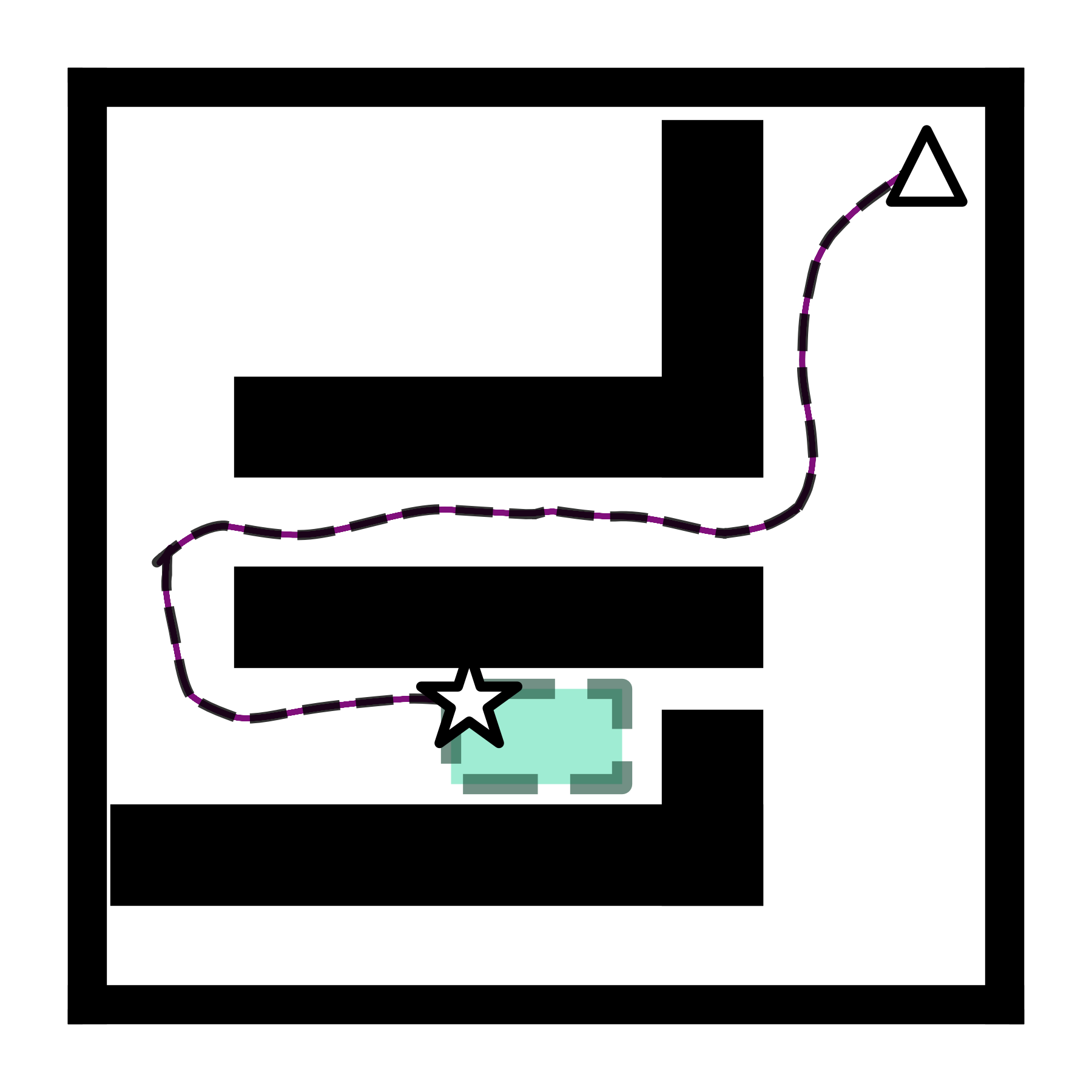}
        \caption{\centering LQRm}
        \label{fig:path_plot_lqrm_0p0000005}
    \end{subfigure}%
    ~
    \begin{subfigure}[t]{0.48\linewidth}
        \centering
        \includegraphics[width=0.99\linewidth]{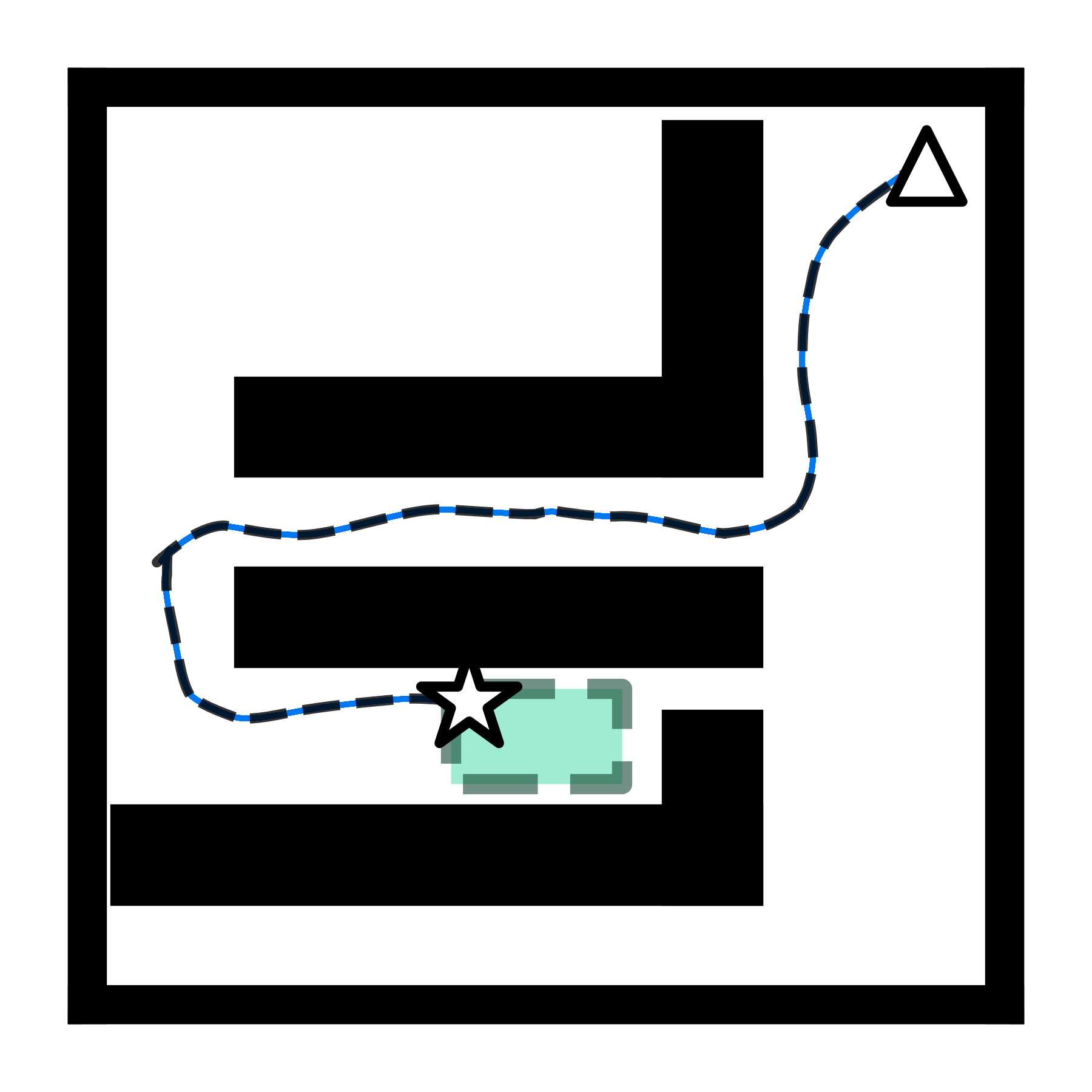}
        \caption{\centering NMPC}
        \label{fig:path_plot_nmpc_0p0000005}
    \end{subfigure}
    \caption{The results of 1000 independent Monte Carlo trials with different low-level reference tracking controllers. The reference trajectory was planned \textbf{with} distributionally robust obstacle padding, and is shown as a dashed line. The goal area is a green rectangular area whose border is marked by a dashed line. The terminal state of failed and successful trajectories are shown by `O' and `X' markers respectively. Obstacles and the free space are represented by solid black and white regions, respectively. The disturbance variance was $\sigma_w^2 = 0.0000005$.}
    \label{fig:path_plot_0p0000005}
\end{figure}

\clearpage

\clearpage

\clearpage

\end{document}